\documentclass[pdflatex,sn-mathphys-num]{sn-jnl} 
\usepackage{graphicx}%
\usepackage{multirow}%
\usepackage{amsmath,amssymb,amsfonts}%
\usepackage{amsthm}%
\usepackage{mathrsfs}%
\usepackage[title]{appendix}%
\usepackage{xcolor}%
\usepackage{textcomp}%
\usepackage{manyfoot}%
\usepackage{booktabs}%
\usepackage{algorithm}%
\usepackage{algorithmicx}%
\usepackage{algpseudocode}%
\usepackage{listings}%

\usepackage{todonotes}
\newcommand{\rev}[1]{\textcolor{black}{#1}}
\newcommand{\revD}[1]{\textcolor{black}{#1}}

\usepackage{textcomp}

\begin{document}

\title[Experimental Validation Marsupial System]{Long Duration Inspection of GNSS-Denied Environments with a Tethered UAV-UGV Marsupial System }

\author[1]{\fnm{Simón} \sur{Martínez-Rozas}}\email{simon.martinez$@$uantof.cl}

\author*[2]{\fnm{David} \sur{Alejo}}\email{dalejo@us.es}

\author[3]{\fnm{José Javier} \sur{Carpio}}\email{jjcarjim@upo.es}

\author[3]{\fnm{Fernando} \sur{Caballero}}\email{fcaballero@upo.es}

\author[3]{\fnm{Luis} \sur{Merino}}\email{lmercab@upo.es}

\affil*[1]{\orgdiv{Department of Electrical Engineering}, \orgname{University of Antofagasta}, \orgaddress{\street{Av. Angamos 601}, \city{Antofagasta}, \postcode{1270300}, \state{Antofagasta}, \country{Chile}}}

\affil[2]{\orgdiv{Department of Systems and Automatic Control Engineering }, \orgname{University of Sevilla}, \orgaddress{\street{Avda. de los Descubrimientos S/N}, \city{Sevilla}, \postcode{41092},  \country{Spain}}}

\affil[3]{\orgdiv{Service Robotics Lab}, \orgname{University Pablo de Olavide}, \orgaddress{\street{Ctra. de Utrera S/N}, \city{Sevilla}, \postcode{41013}, \country{Spain}}}

\maketitle

\abstract{Unmanned Aerial Vehicles (UAVs) have become essential tools in inspection and emergency response operations due to their high maneuverability and ability to access hard-to-reach areas. However, their limited battery life significantly restricts their use in long-duration missions. This paper presents a tethered marsupial robotic system composed of a UAV and an Unmanned Ground Vehicle (UGV), specifically designed for autonomous, long-duration inspection tasks in Global Navigation Satellite System (GNSS)-denied environments. The system extends the UAV’s operational time by supplying power through a tether connected to high-capacity battery packs carried by the UGV. Our work details the hardware architecture based on off-the-shelf components to ensure replicability and describes our full-stack software framework used by the system, which is composed of open-source components and built upon the Robot Operating System (ROS). The proposed software architecture enables precise localization using a Direct LiDAR Localization (DLL) method and ensures safe path planning and coordinated trajectory tracking for the integrated UGV–tether–UAV system. We validate the system through \revD{three sets of field experiments involving (i) three manual flight endurance tests to estimate the operational duration, (ii) three experiments for validating the localization and the trajectory tracking systems, and (iii) three executions of an inspection mission to demonstrate autonomous inspection capabilities.} The results of the experiments confirm the robustness and autonomy of the system in GNSS-denied environments. Finally, all experimental data have been made publicly available to support reproducibility and to serve as a common open dataset for benchmarking.\\\\
}

% Keywords
%\keyword{GNSS-denied areas; Tethered robots; long-duration inspection; marsupial system} 

% The fields PACS, MSC, and JEL may be left empty or commented out if not applicable
%\PACS{J0101}
%\MSC{}
%\JEL{}

%%%%%%%%%%%%%%%%%%%%%%%%%%%%%%%%%%%%%%%%%%

%%%%%%%%%%%%%%%%%%%%%%%%%%%%%%%%%%%%%%%%%%

\section{Introduction}

\label{sec:introduction}

Research and deployment of Unmanned Aerial Vehicles (UAVs) have experienced exponential growth during the 21st century in a broad variety of civil applications \cite{NEX2022215}. 
Some applications that include UAVs as the main agents are road traffic monitoring, remote sensing, search and rescue operations in emergency situations, security and surveillance over an area, assessment and spraying in agriculture \cite{info10110349}, forest firefighting \revD{\cite{viegas_2022_tethered_firefighting}}. In them, UAVs are widely used due to their high maneuverability, mobility, and the decreasing costs of both purchase and maintenance.  Other applications that have benefited from the use of UAVs are related to inspection tasks to analyze and monitor civil infrastructure \cite{ietrsn20170251,101007978981197331452}. These tasks use the capabilities of the UAV to move within closed, dangerous, or risky environments or to reach objectives that are at a great distance. For example, UAVs have been used in the evaluation of the structure of buildings and bridges \cite{sreenath_2020}, the inspection of power lines, the monitoring and extinguishing of fires in the facade and interior of buildings \revD{\cite{viegas_2022_tethered_firefighting}}.%, and even the inspection of sewers \cite{jfralejo}.

However, a significant limitation of small and medium-sized UAVs is their flight duration, which hinders their ability to conduct comprehensive inspection tasks. In fact, the majority of commercial platforms exhibit flight times of less than thirty minutes, typically measured under laboratory conditions and without any additional payload, such as computing devices or sensors \cite{10623695}. \revD{Fully equipped autonomous drones carrying the necessary sensor payload seldom achieve flight times of more than twenty minutes \cite{drones7070471}. }

There are some alternatives \revD{to develop long-duration inspection tasks}. The most common approach is to perform battery swapping either manually, thus requiring human intervention, or automatically with battery swapping stations \cite{8204465}, requiring the installation of additional infrastructure. Another alternative \revD{is to use a team of robots flying either alternately or simultaneously, so that the inspection mission can be carried out without interruptions or in a shorter amount of time \cite{KIM2018291}. However, the use of a team of UAVs greatly increases the complexity of the system and its acquisition and maintenance costs. Finally, tethered UAVs connected to a base station that supplies them with power have attracted the attention of the research community} \cite{tethered_survey, drones9060425} with applications as diverse as collaborative load transportation \cite{9213842}, high-bandwidth communications \cite{850822,9202196} and airborne wind energy \cite{7857772}. \revD{Regrettably, the use of a fixed tether at a specific location significantly limits the operational range of the UAV, thereby preventing safe access to certain areas of interest.}

In this paper, we present an autonomous tethered marsupial robotic configuration for long-duration inspection tasks. Our system utilizes an Unmanned Ground Vehicle (UGV) to transport battery packs that supply power through a cable to a UAV equipped with the sensors needed to carry out the inspection. 
This approach is based on a philosophy of robotic collaboration, offering remarkable advantages over traditional standalone systems by leveraging the unique strengths of each robotic agent \cite{MooreIROS2018}. A case of success of a marsupial robotic system can be found in the Defense Advanced 
Research Projects Agency (DARPA) 2021 Subterranean Challenge \cite{DARPA2021}, where the CSIRO Data61 Team reached second place \cite{CSIRO2024}. The team used a marsupial system in which a UGV transports a UAV to a take-off point to carry out inspection tasks, which reduces the UAV's energy consumption. Another successful case is the Mars 2020 rover and helicopter \cite{Grip2018GuidanceAC} which are being used for planetary exploration on Mars. In this system, the helicopter augments the capabilities of the rover, exploring large areas faster than the rover and providing reconnaissance on target locations and safe-to-traverse routes. In both applications, the concept of collaboration based on the robotic strengths is used; however, the robots are not tethered to provide energy, so the UAV operation time is still limited by the power it can carry. 

\revD{Following the tethered marsupial approach, an important milestone is introduced in \cite{oxpecter_drones7020073},} where the authors describe the development of a tethered UAV system operating alongside a UGV for inspecting stone-mine pillars and assessing structural integrity in underground mines. The UAV is powered via a tether from the UGV, allowing for extended operation without relying solely on its onboard battery. The UGV carries additional batteries, providing continuous power through the tether and effectively addressing the UAV’s flight-time limitations. Despite the robustness of the implemented system, the work focuses on autonomous navigation and inspection around stone pillars, rather than enabling navigation and inspection in general or more complex environments. Furthermore, the article does not provide an analysis of the UAV’s long-term inspection capabilities. A similar collaborative UAV-UGV system connected by a power tether is presented in \cite{6961531}. This system leverages the strengths of both platforms to extend autonomous navigation in partially mapped environments: the UGV, equipped with a large power supply, powers the UAV, while the UAV contributes aerial surveillance, enhanced situational awareness, and obstacle detection for the UGV. However, this work lacks a detailed description of the system architecture and does not analyze flight duration.

Another approach based on a marsupial system is presented in \cite{capitán2024maspaefficientstrategypath}, where the authors introduce an optimal path-planning strategy for a marsupial robotic system consisting of a UGV, a UAV, and a tether connecting both robots. The system follows a sequential navigation strategy through 3D environments, avoiding both ground and aerial obstacles to ensure a collision-free trajectory for the team. Nonetheless, this work does not present a complete analysis of the system’s suitability for autonomous inspection tasks or evaluate its flight endurance. In \cite{gnss-denied-gross-2019}, the authors introduce a collaborative UGV-UAV system for reliable localization and autonomous operation in Global Navigation Satellite System (GNSS)-denied environments. The UGV, equipped with a 3D LiDAR, a fisheye camera, an Ultra-WideBand (UWB) radio, and an Inertial Measurement Unit (IMU), handles mapping and localization, while the UAV uses shared IMU and laser altimeter data from the UGV. An Error-State Extended Kalman Filter fuses this sensor data to estimate the UAV’s relative position, enabling real-time tracking and 3D mapping. However, the system does not consider the tether in its planning process, nor does it address UAV flight-time limitations. In \cite{maese2025physicalsimulation}, a simulator based on the Robotic Operating System (ROS) version 2 and the Gazebo simulator is presented for marsupial systems where a UAV and a UGV are connected by an adjustable-length hanging tether. This configuration enables realistic modeling of the interactions between the robots and the winch that regulates \revD{the length of the tether}. Its primary contribution is to offer a tool for validating and testing planning and control algorithms, thereby reducing the costs and risks of physical experiments. However, the work is limited to serving as a tool for testing navigation modules, and it does not facilitate field experiments to quantify the viability of this type of system, the UAV’s flight duration, or even the battery charge status.

\begin{figure}[H]
 %   \centering
    \begin{tabular}{cc}
    
    {\includegraphics[width=0.46\linewidth]{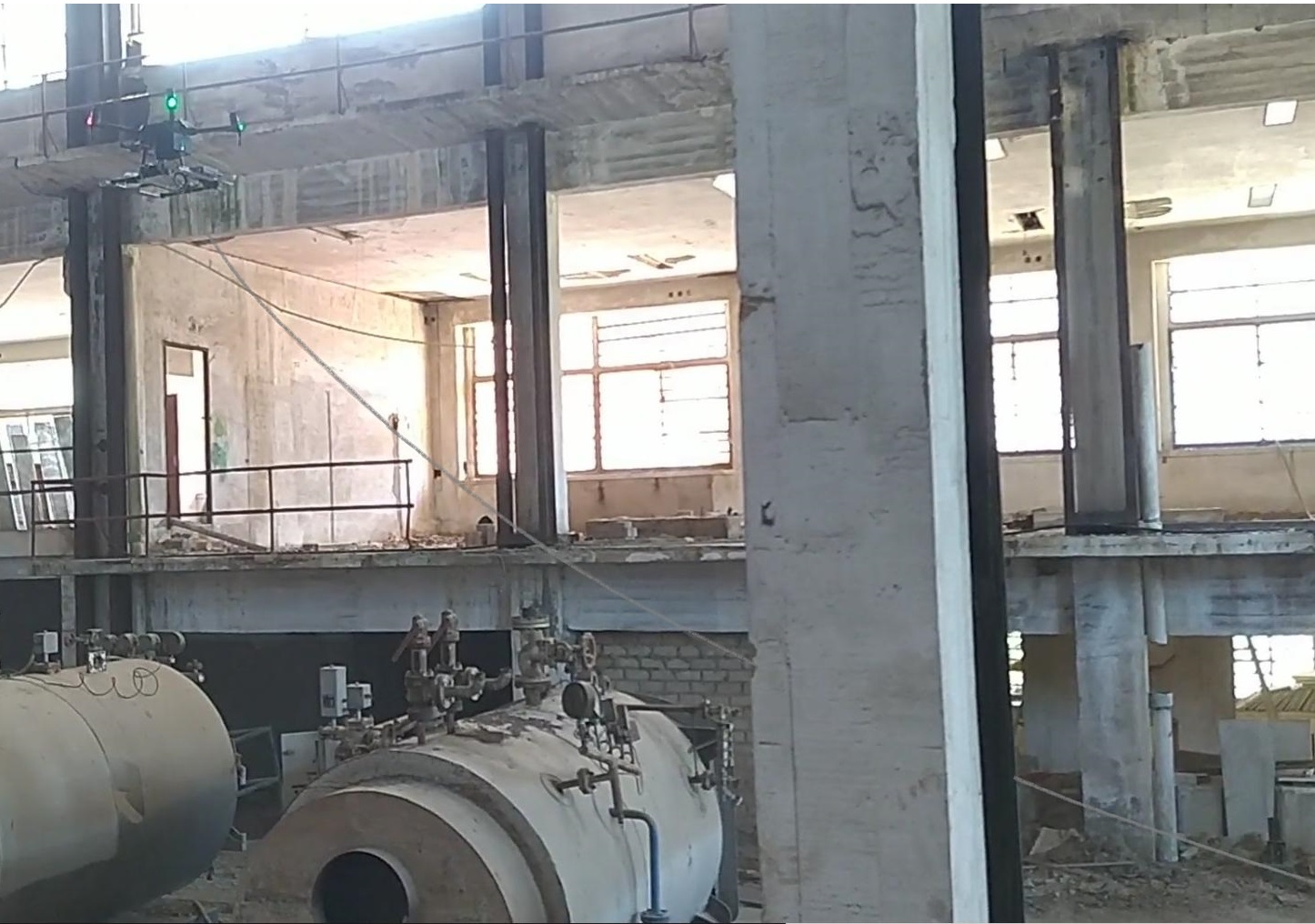}} &
    {\includegraphics[width=0.49\linewidth]{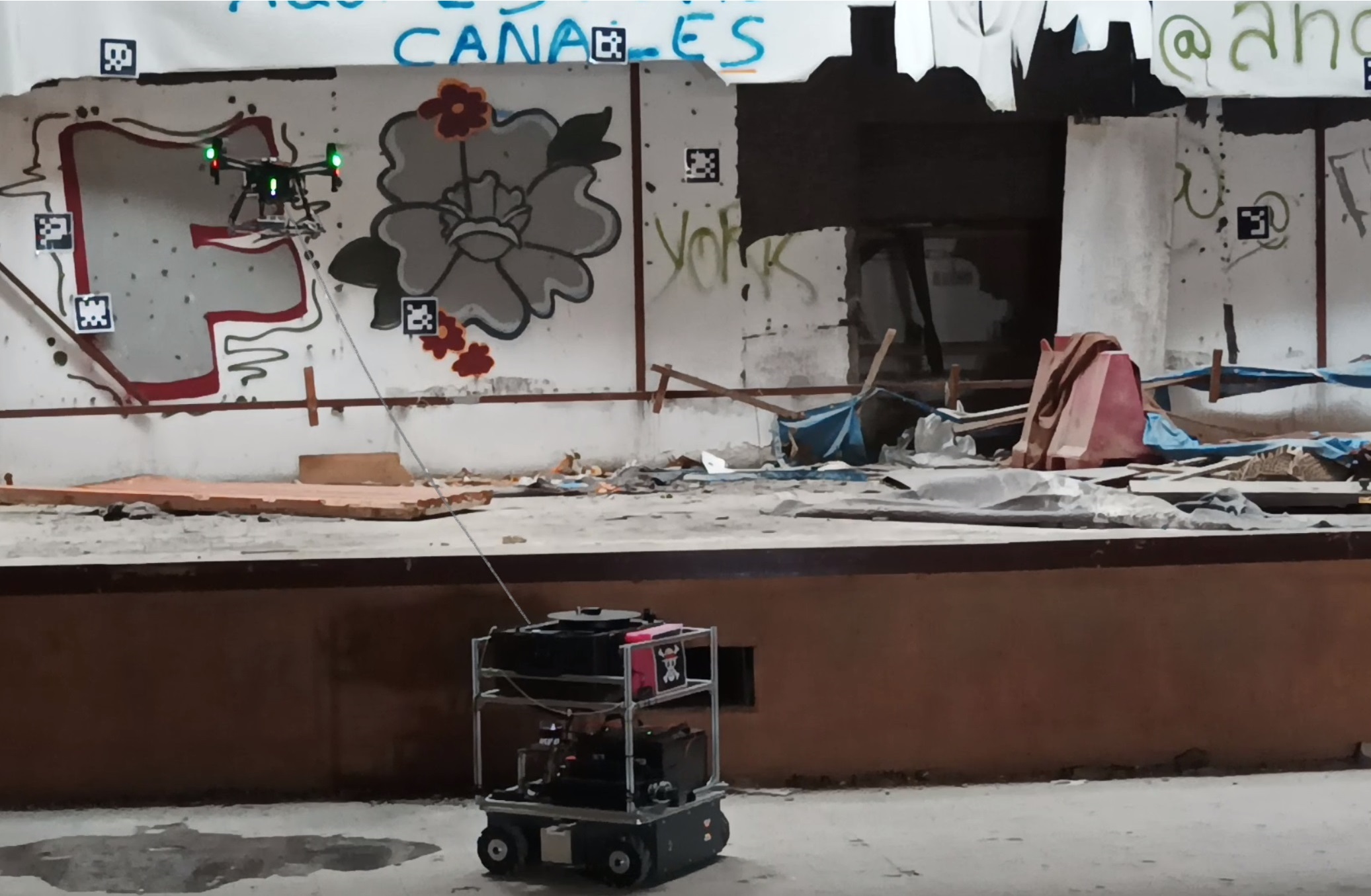}} \\
     (\textbf{a}) & (\textbf{b})
    
    \end{tabular}
    \caption{Our marsupial system performing autonomous inspection tasks in two buildings with structural damage in the University Pablo de Olavide (UPO), Seville (Spain). (\textbf{a}) Detail of the UAV at the abandoned thermal station (Scenario 2). (\textbf{b}) Tethered marsupial system   at the old theater building (Scenario 3).}
    \label{fig:system_proposed}
    
\end{figure}

This paper extends our previous work presented in \cite{10557064}, in which we presented a short description of the proposed system for long-duration missions. As an extension, this paper gives a detailed description of the main onboard software modules used for control, localization and navigation purposes. \rev{Moreover, we present here extended field reports that show that our proposed system can  perform autonomous inspection tasks in cluttered environments}, which are still a challenge in tethered UAV-UGV systems \cite{drones9060425}. To this end, we present a set of validation experiments, including quantitative localization and navigation results. Thus, the main contributions of this paper are as follows:

\begin{enumerate}
    \item  \revD{We present the hardware and software design of the proposed marsupial robotic system as a whole (see Figure \ref{fig:system_proposed}). It consists of a UGV linked to a UAV and is built using off-the-shelf components } in contrast to existing approaches in the literature \cite{oxpecter_drones7020073}.  In our system, we use a \revD{cable} that connects the UAV to the UGV. This \revD{tether supplies the UAV with power}, enabling long-term inspection missions. \revD{To the best of our knowledge, this is the first system to demonstrate a continuous operation exceeding two hours.}
    \item We present the architecture of our software solution. The different software modules, developed in-house, are released as open-source. All the modules have been incrementally validated to reach the final goal of performing autonomous inspection missions with a tethered UAV-UGV marsupial system. This is a significant contribution with respect to \cite{10557064,dll,smartinezr2023}. 
    \item  We pay special emphasis on describing our developed localization and navigation systems. In particular, the localization system is based on the one presented in our previous work \cite{dll}, but is adapted for the current case and for providing multi-robot localization in Section \ref{sec:localization}. Regarding navigation, we present a reformulation of our trajectory planner from \cite{smartinezr2023}, adapted to account for the constraints of the proposed marsupial system. In particular,  in this paper we always consider a \revD{straight} tether instead of the variable-length hanging tether approach presented in \cite{smartinezr2023}.
\end{enumerate}

The paper is structured as follows: Section \ref{sec:overview} provides a basic description of the hardware of the UGV, the UAV and the tether systems, including details on the power system for the UAV. It also presents the basic software architecture of the system. Then, we describe in detail the localization and navigation systems used in autonomous inspection missions in Sections \ref{sec:localization} and \ref{sec:navigation}, respectively. Our marsupial system has been extensively validated in experimental tests in three different scenarios. These results are presented in Section \ref{sec:experiments}. Later, we present an extended discussion of our results in Section \ref{sec:discussion}. Finally, the conclusions and future research directions are detailed in Section \ref{sec:conclusions}.

%%%%%%%%%%%%%%%%%%%%%%%%%%%%%%%%%%%%%%%%%%
\section{Materials and Methods}
\label{sec:overview}

\begin{figure}[H]
    %\centering
    \begin{tabular}{cc}
    \includegraphics[height=0.25\textheight]{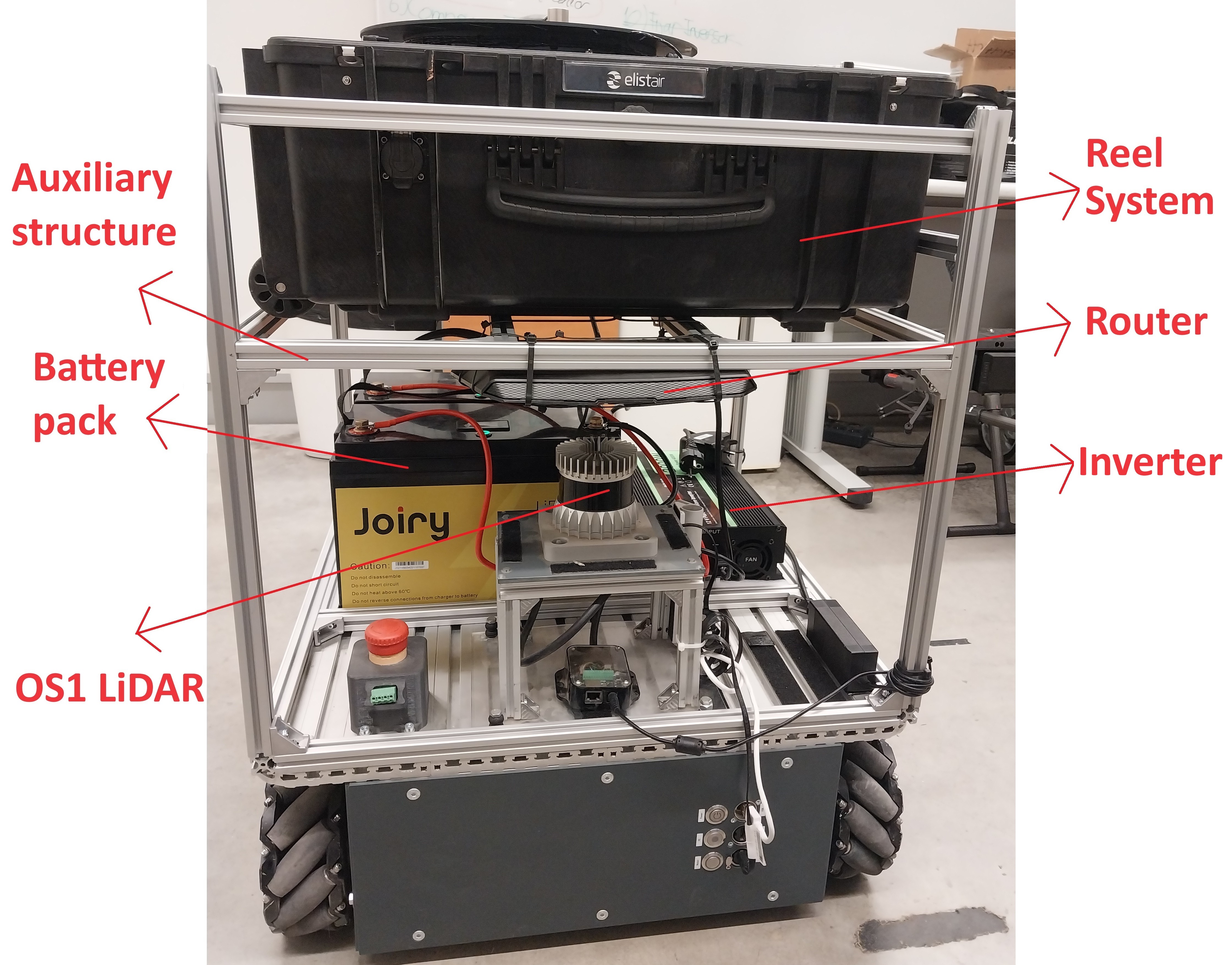} & \includegraphics[height=0.25\textheight]{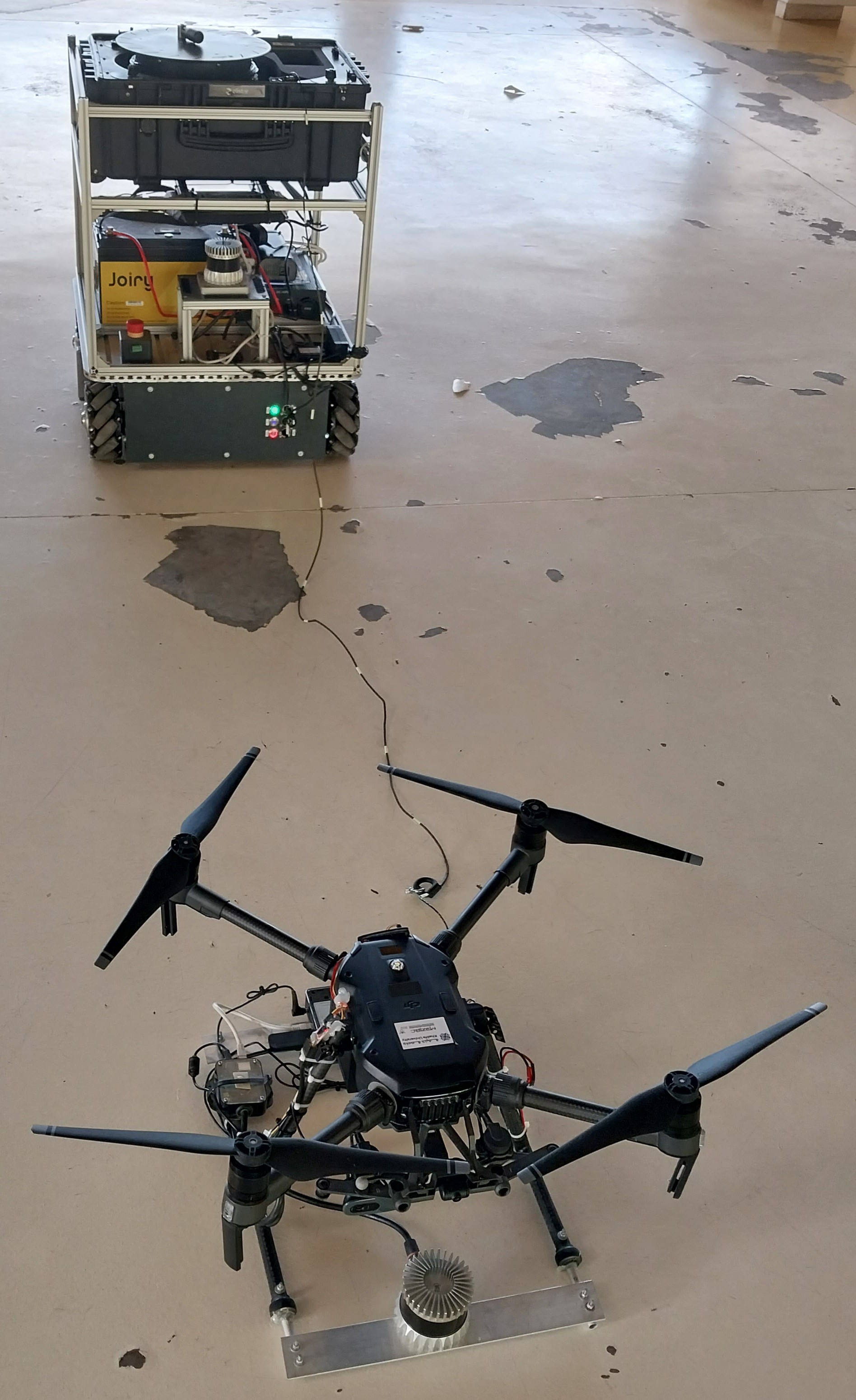} \\
    (\textbf{a}) & (\textbf{b})
    \end{tabular}
    \caption{Marsupial system, UAV tied to UGV. (\textbf{a}) Detail of the UGV equipped with the auxiliary structure that stores the systems of the marsupial configuration. (\textbf{b})  Marsupial system in an initial position before the experiment.}
    \label{fig:system_complete}
\end{figure}

In this section, we describe the main \revD{components} of our marsupial system. The system consists of a UGV, a UAV, and a tether system (see Figure \ref{fig:system_complete}), which \revD{supplies} the UAV with power from additional battery packs onboard the UGV. \revD{This configuration allows us to significantly} extend the flight duration of the UAV. The UGV, the UAV and the tether system are described in Sections \ref{sec:ugv_section}, \ref{sec:uav} and \ref{sec:tether_section}, respectively. Section \ref{sec:power} analyzes the power requirements \revD{used to determine} the capacity of the batteries on the UGV. Finally, Section \ref{sec:software} gives an overview of our software solution for autonomous operation.

\subsection{Unmanned Ground Vehicle (UGV)} 
\label{sec:ugv_section}

\begin{figure}[H]
   % \centering
    \includegraphics[width=0.5\linewidth]{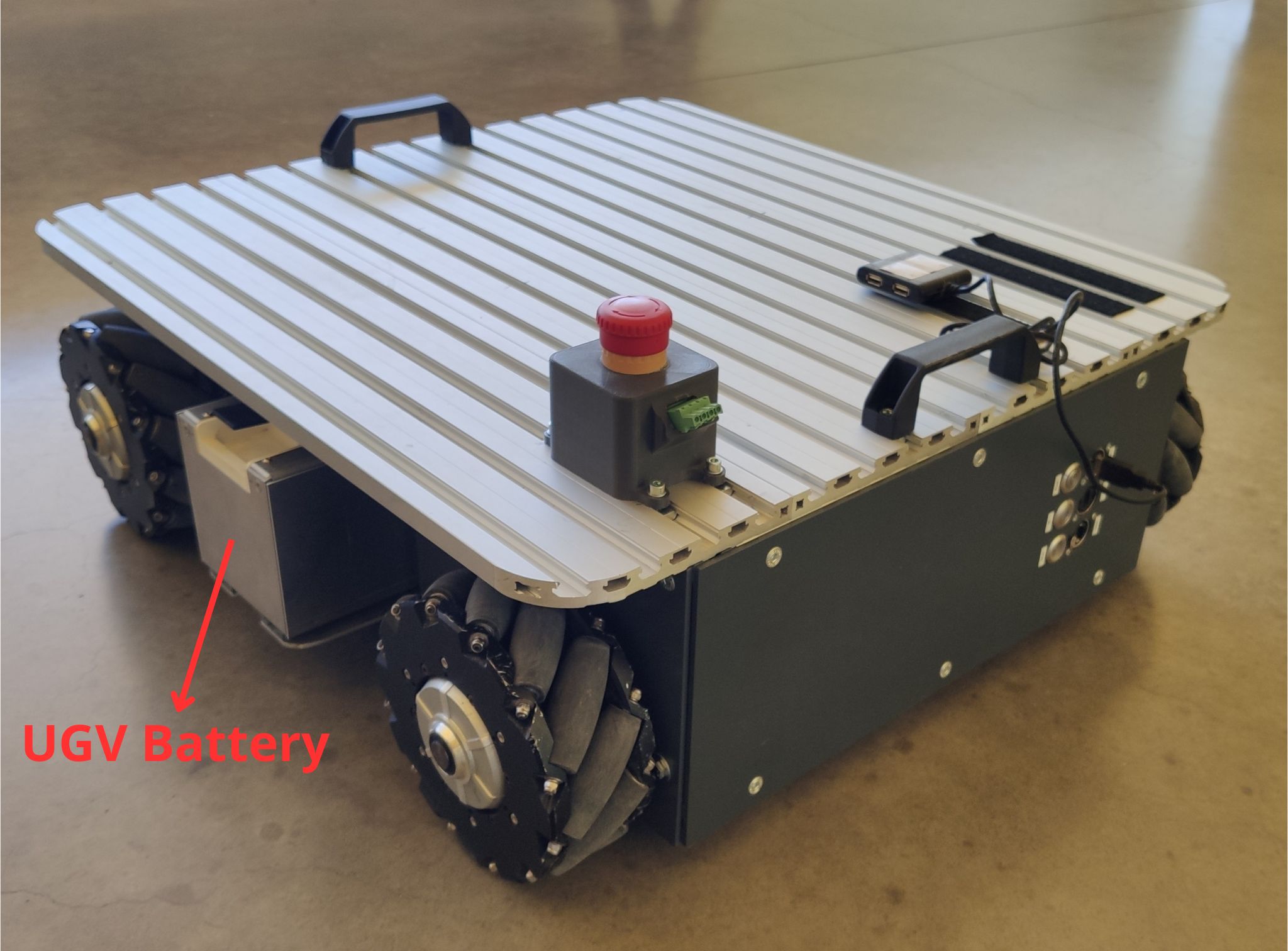}
    \caption{Basic ARCO UGV from IDMind with its onboard batteries.}
    \label{fig:arco}
\end{figure}

We use an ARCO ground platform from IDMind, which is a four-wheeled holonomic robot with an independent traction system designed for the delivery of heavy loads in factories (see Figure \ref{fig:arco}). The base platform has two batteries for motors and electronics, respectively, that give ARCO an autonomy of three hours of operation with a maximum payload of over one hundred kilograms and a maximum speed of 0.8 m/s.

Additionally, we have properly equipped ARCO with the necessary components to achieve the marsupial configuration (see Figure \ref{fig:system_complete}a). Most importantly, the ARCO platform \revD{includes an internal computer that  runs the internal software responsible for processing data from its IMU and 3D LiDAR sensors to localize the platform in real time.}  It also provides the rest of the system with \revD{high-speed 5G Wi-Fi connectivity via its integrated router. Finally, the tether system and additional batteries for powering the UAV are installed on the platform. All components} of the ARCO platform are listed in Table \ref{tab:arco_components}.

\begin{table}[ht] 

\caption{Main components of the ARCO UGV platform.}
\label{tab:arco_components}

\begin{tabular}{llc}

      \textbf{Component} & \textbf{Model} & \textbf{Quantity}\\

Onboard Computer & MSI Cubin 8GL-001BEU (i5-11400) & 1 \\ 
     IMU & Sparkfun Razor 9 DoF & 1\\
     LiDAR & Ouster OS1-16 & 1\\
     UGV Internal Battery &    LiFePO 75 Ah@24 V & 2\\
     UAV Power Battery & Joiry LiFePO 150 Ah@12 V & 2 \\
     Inverter & Green Cell 2000 W/4000 W &1 \\
     Tether System & Elistair LIGH-T V4 Tethered Station (LTS4)&1 \\
     WiFi Router & NIGHTHAWK NetGear AX5400 &1 

\end{tabular}

\end{table}

\subsection{Unmanned Aerial Vehicle (UAV)} 
\label{sec:uav}

\begin{figure}[H]
  %  \centering
    \includegraphics[width=0.5\linewidth]{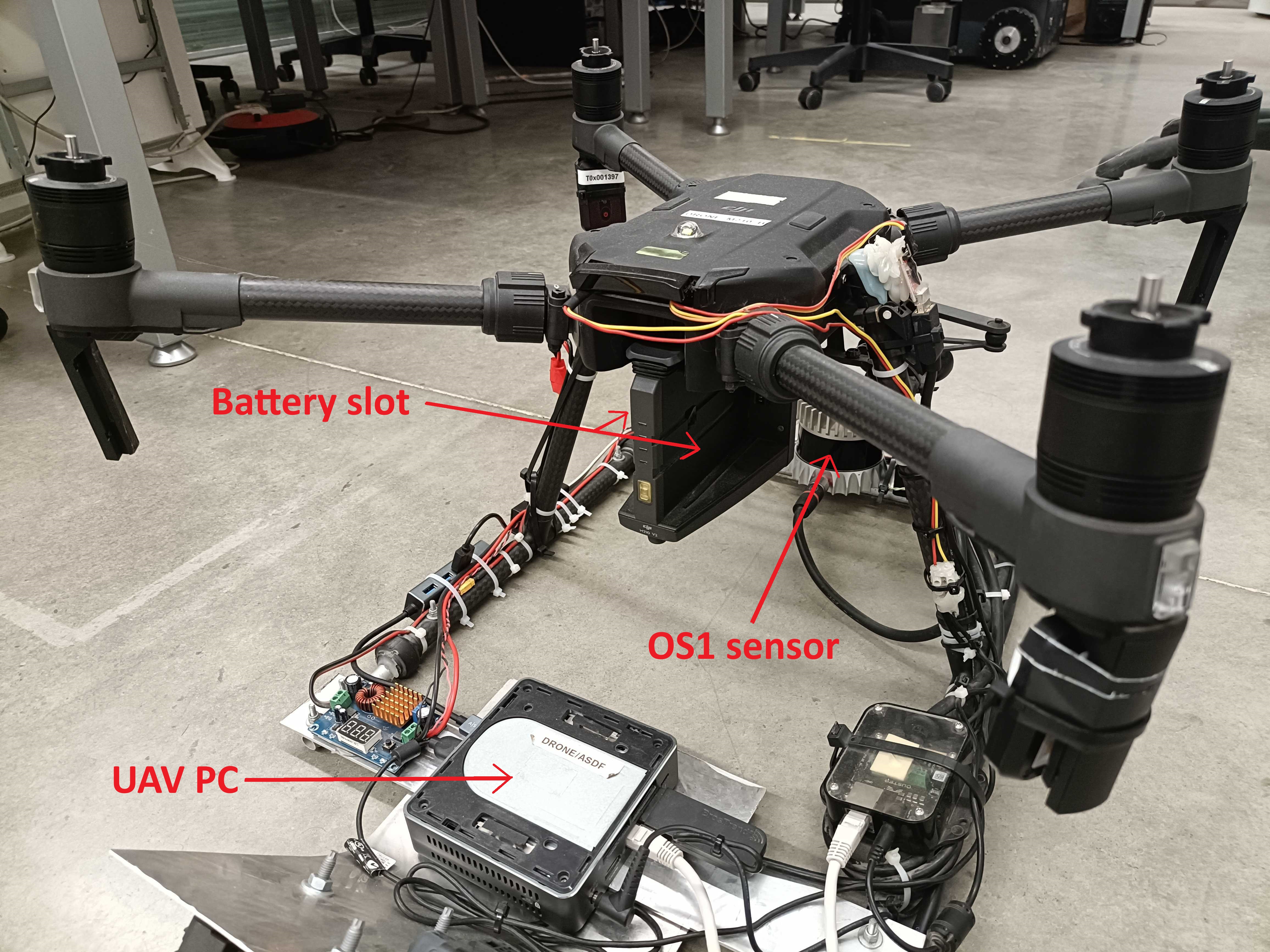}
    \caption{M210 UAV and its additional onboard systems.}
    \label{fig:dji_platform}
\end{figure}

Regarding the aerial platform, the model used for the experiments is the DJI Matrice 210 V2 (M210) (see Figure \ref{fig:dji_platform}). We selected this platform because of the robustness of the hardware and software, the support that the provider offers, and its ROS integration by means of a Software Development Kit (SDK). 

The main additions are its onboard computer and a LiDAR sensor, as shown in Figure~\ref{fig:dji_platform}. In addition, we have installed an external access point to increase the throughput and robustness of the WiFi connectivity (see Table \ref{tab:uav_components}).

\begin{table}[t!]
\caption{Main components added to the UAV platform}
\label{tab:uav_components}

\begin{tabular}{|c|c|} \hline
     \textbf{Component} & \textbf{Model}   \\ \hline
     Onboard computer & NUC8i7BEH 32GB RAM  \\
     LiDAR & Ouster OS1-16 \\
     Internal battery & DJI TB55 (backup) \\
     WiFi Access Point & D-Link AX1800 Wi-Fi USB Adapter \\ \hline
\end{tabular}

\end{table}

\subsection{Tether System}
\label{sec:tether_section}
\begin{figure}[H]
    \centering
    \includegraphics[width=1.0\linewidth]{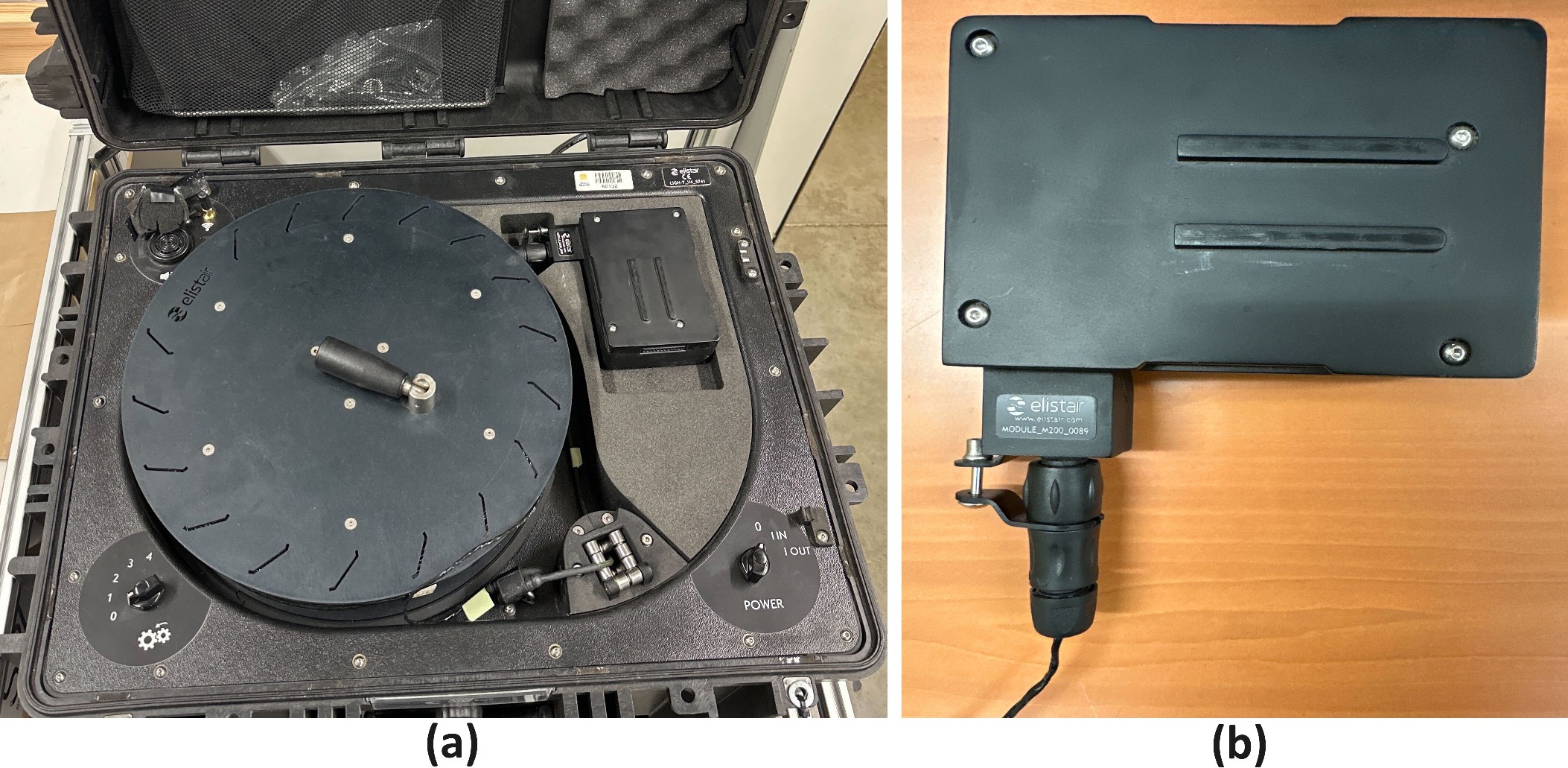}
    
    \caption{(\textbf{a}) LIGH-T V4 Tethered Station (LTS4), including switches, a reel for the cable, and the battery connector. (\textbf{b}) Detail of the battery connector. }
    \label{fig:elistair}
\end{figure}

We use a LIGH-T V4 Tethered Station (LTS4), provided by the company Elistair \cite{elistair_web} (see Figure \ref{fig:elistair}) to power the UAV, due to its compatibility with the M210, among other DJI models. The cable of the system ends in an adapter, which is located in one of the two drone battery slots to provide energy to the drone. The maximum length of the \revD{cable} is around seventy meters. Note that a fully charged battery has to be equipped in the UAV in the remaining battery slot on the M210. This battery acts as a backup that is used in case of a disconnection between the UAV and the LTS4. \revD{The cable is connected to a DC motor, which can perform a constant tension on the cable ranging from one to three Newtons.}

\subsection{Power System Requirements}

\label{sec:power}

\begin{figure}[H]
    %\centering
    \includegraphics[width=1.0\linewidth]{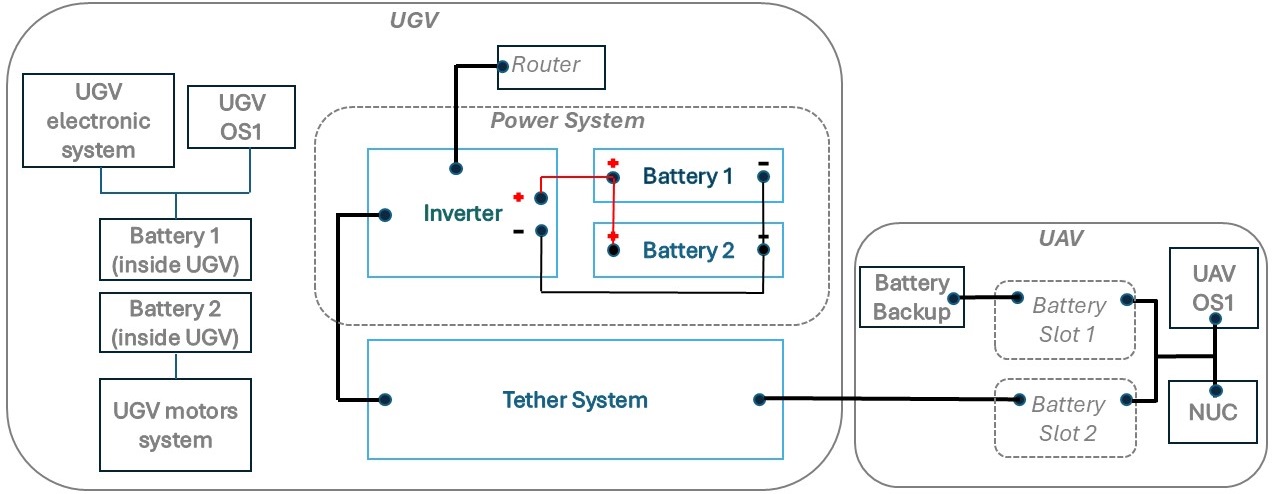}
    \caption{Connection diagram of the power system onboard the UGV and the UAV.}
    \label{fig:conection_diagram}
\end{figure}

We have designed our power system to allow the UAV to fly up to two hours with a remaining battery level above 10\%. Figure \ref{fig:conection_diagram} represents the interconnection scheme of the power system, which consists of the following components:

\begin{itemize}    
    \item We have installed a compact Green Cell® 2000 W/4000 W inverter from 12 V DC to 220 V/230 V AC, \revD{with an effectiveness of 85\%}. This inverter powers the LTS4 and the \revD{internal UGV router} during the experiment.

    \item We use the Joiry LiFePO4 batteries, which have a capacity of 150 Ah at a nominal voltage of 12.8 V. With one battery we obtain a theoretical flight duration \revD{slightly over seventy minutes, taking into account the effectiveness of the inverter}. Thus, we connect two batteries in parallel to further extend the flight duration (see Figure~\ref{fig:conection_diagram}). LiFePO4 batteries have great characteristics, including high energy density, high current discharge, and long durability. Most importantly, they are safer to operate when compared to LiPO batteries \cite{ZHU2021210564}. 

    \item A backup battery is connected to the UAV in the remaining battery slot of the M210. We use the battery model TB55 that provides the UAV with about ten minutes of flight duration in case of failure of the LTS4. This time is sufficient for the monitoring operator to take control of the UAV and land it safely in the event of an LTS4 system malfunction.
    
\end{itemize}

\subsection{Software Architecture}
\label{sec:software}

\begin{figure}[H]
   % \centering
    \includegraphics[width=1.0\linewidth]{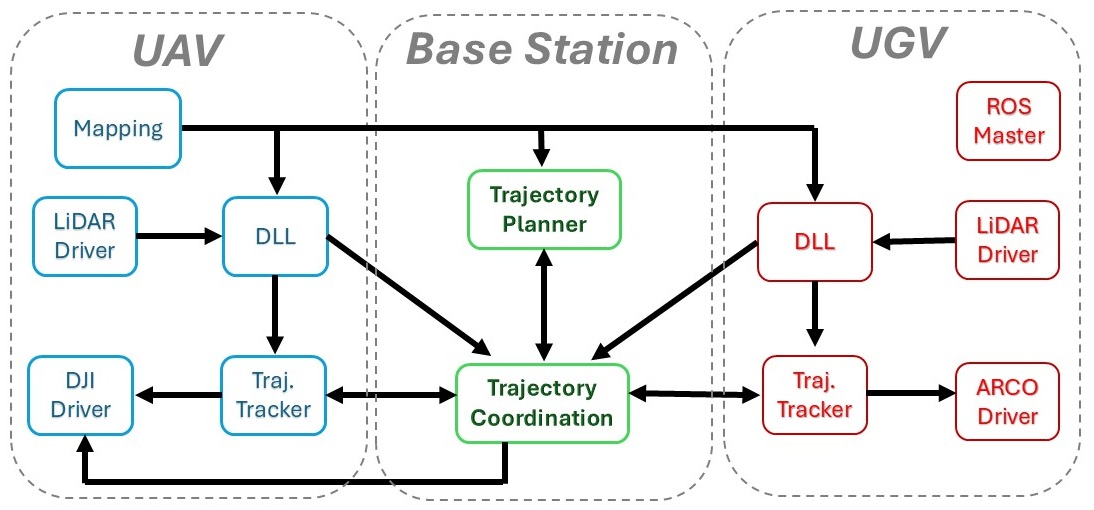}
    \caption{Main ROS nodes of the marsupial system used to execute autonomous missions are distributed across the UAV computer (in blue), the Base Station (in green), and the UGV (in red). The arrows indicate the flow of information and its predominant direction. The ROS master node is required only to establish communication between the nodes.}
    \label{fig:software}
\end{figure}

\revD{The proposed marsupial system relies on a distributed software architecture implemented entirely in ROS Noetic running on Ubuntu 20.04. ROS provides the natural framework for decomposing the system into independent modules, each implemented as a ROS node, which can run on the UGV, the UAV, or the laptop used as the base station. Figure~\ref{fig:software} illustrates the overall architecture, showing the nodes allocated to each platform. The functionality of these nodes is described in detail in the following sections. All software modules are openly available through the GitHub profile of the Service Robotics Lab, Universidad Pablo de Olavide (UPO) (
\url{https://github.com/robotics-upo}, accessed on 3 Nov 2025), ensuring reproducibility and transparency.}

\revD{At the base station, two key modules orchestrate the system. The Trajectory Planner employs a two-step procedure adapted from \cite{smartinezr2023}: first, an initial solution is generated with a Rapidly exploring Random Tree Star (RRT*), and then refined using a non-linear optimizer that incorporates the kinematic and dynamic constraints of both robots while improving safety margins. The second module, the Trajectory Coordinator Module (TCM), ensures the synchronized motion of the UAV and UGV during joint missions.}

\revD{Onboard the UGV, the architecture begins with the ROS Master, which establishes the communication backbone for all distributed nodes. The robot’s hardware is accessed through the ARCO driver, which exposes standard ROS commands and services. Environmental perception is provided by the 3D LiDAR driver, whose data feeds directly into the Direct LiDAR Localization (DLL) module \cite{dll}, responsible for localization against a prior 3D map (see Section \ref{sec:dll} for more details). Trajectory execution is managed by the Trajectory Tracker, which translates planned paths into actionable commands for the ARCO driver (see Section \ref{sec:trajectory_tracking} for more details).}

\revD{The UAV runs a similar software stack, with specific adaptations for aerial operation. The DJI driver interfaces with the UAV’s onboard hardware, exposing sensors and actuators through ROS topics and services. Like the UGV, the UAV also runs the 3D LiDAR driver and the DLL module, ensuring accurate positioning even in GNSS-denied environments. Finally, the UAV executes trajectories using its own trajectory tracker, similar in functionality to the UGV’s tracker but extended with a proportional controller for altitude regulation.}

\revD{Altogether, this modular and distributed software architecture enables robust cooperation between the UAV and UGV, leveraging open-source ROS modules to provide a reproducible framework for marsupial robotics research.}

\section{Localization System}
\label{sec:localization}

Having a precise localization of the marsupial system is crucial to ensure that the inspection is performed as expected during the mission. In outdoor scenarios with good GNSS coverage, we can use Differential or Real-Time Kinematic (RTK) approaches to obtain a precise localization estimation with errors of a few centimeters. However, performing close inspections of large objects (facades, large statues, buildings, etc.) based on GNSS might be subject to significant localization errors (up to several meters) due to the reduction in GNSS satellite visibility and the consequent poor Dilution of Precision (DOP) \cite{8403395}. LiDAR-based localization approaches have been demonstrated to be reliable and accurate enough to perform long-term robot localization, both aerial and terrestrial. 

In this paper, we focus on the localization problem in GNSS-denied areas of our marsupial system. To this end, we use a two-step approach. First, Section \ref{sec:mapping} describes the mapping stage, when we obtain a 3D map of the environment by using a LiDAR sensor. Then, this map is used with the localization method described in Section \ref{sec:dll} when performing the inspection mission using the same LiDAR sensor.

\subsection{Mapping Stage}
\label{sec:mapping}
The goal of this stage is to obtain a 3D model of the environment. To this end, a brief manual flight is executed, typically lasting two to five minutes, depending on the spatial extent and complexity of the operational environment. It is important to gather data from all sensors and at different altitudes to ensure proper mapping in the whole scenario. Note that this stage could be omitted as long as a 3D description of the building by means of a CAD file or a Building Information Modeling (BIM) is available, which was not the case in our experimental scenarios.

The purpose of the map is twofold. First, it is used by the localization algorithm as a reference to estimate the pose of both platforms in real time. Second, the 3D map is used to design the inspection plan that should be followed in the experiment. In this paper, we define an inspection plan as a sequence of Points of Interest (PoIs) with orientation that should be visited by the UAV.

\begin{figure}[H]
  %  \centering
    \includegraphics[width=1.0\linewidth]{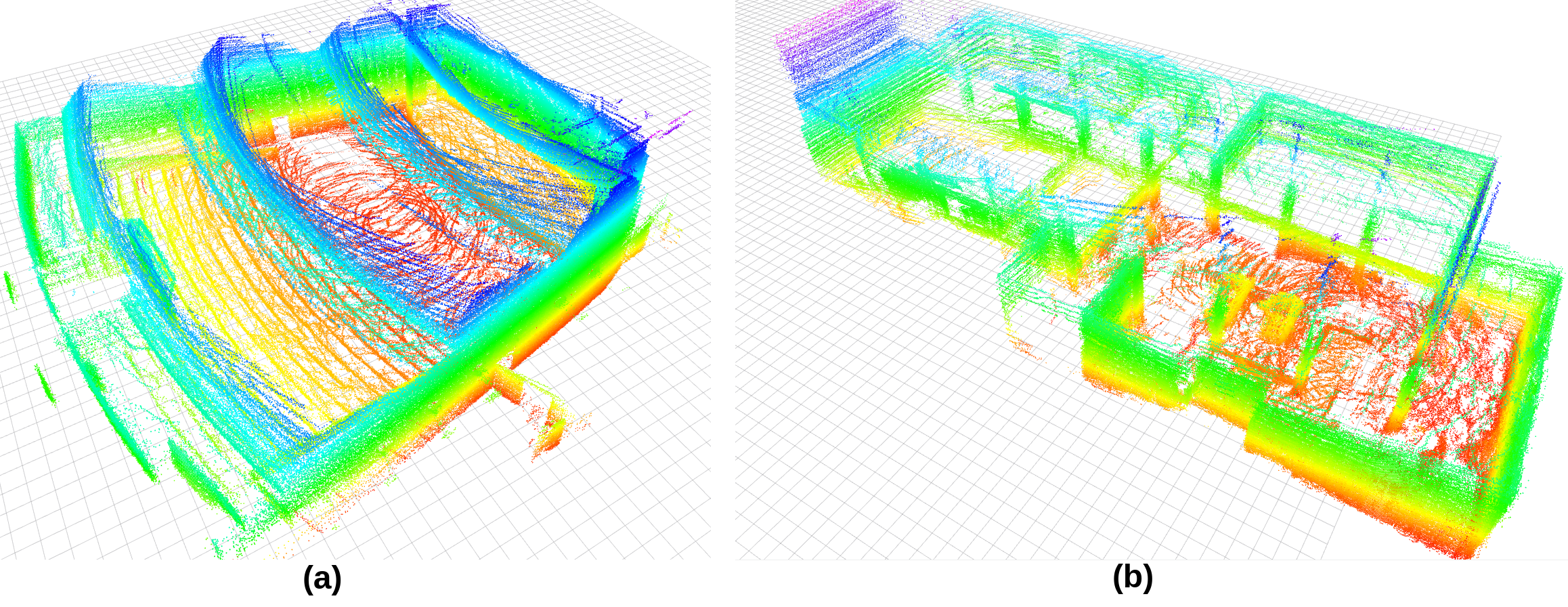} 
    \caption{Point clouds of the mapping process using the A-LOAM package. The color represents the height (z-coordinate) of each point. (\textbf{a}) Theater scenario. (\textbf{b}) Thermal central scenario.}
    \label{fig:maps_example}
\end{figure}

We can use any of the state-of-the-art LiDAR-based odometry and Simultaneous Localization and Mapping (SLAM) algorithms, such as LiDAR Odometry and Mapping (LOAM)~\cite{Zhang-2014-7903} or Versatile and Efficient LiDAR-Inertial Odometry and Mapping (VE-LIOM)~\cite{rs16152772}, to name a few, for experimentally obtaining the 3D map from a sequence of LiDAR and IMU measurements. Our current implementation uses the open-source advanced implementation LOAM (A-LOAM) ROS package \cite{aloam_github}, which is an advanced implementation of the LOAM algorithm that uses Eigen and Ceres Solver to simplify the structure of the software. The mapping results obtained with this package are shown in Figure \ref{fig:maps_example}.

\subsection{Online Localization}
\label{sec:dll}

In this section, we assume that we have a 3D point-cloud of the environment, which can be obtained with the methods specified in Section \ref{sec:mapping}, and we want to have an online estimation of the pose of our vehicles during the experiment so that the inspection mission is properly executed.

\revD{We have designed} an optimization-based localization system, DLL \citep{dll}. It is a fast, direct, map-based localization technique using 3D LiDAR for its application. DLL implements a point cloud to map registration based on non-linear optimization of the distance of the points and the map, thus not requiring features nor point correspondences. Instead, it uses a trilinear interpolation of the Euclidean Distance Field (EDF) representing the environment, which enables enhanced search for the optimal thanks to the continuous gradient computation. The EDF is processed offline during map building, although it can also be generated online with faster implementations like FIESTA \cite{Han:Iros19}, significantly reducing computational load compared to kd-tree-based approaches. 

This method uses a given initial pose in the 3D map. From it, it is able to estimate the pose of the robot by finding the pose that optimally matches the sensed obstacles in the environment to the prior 3D map. To perform the optimization, it uses the translation estimated by the odometry as the initial guess for the non-linear solver. In particular, we make use of a simple sensor integration for short-term odometry for the UAV in order to save computation time. Regarding the UGV, we use fused odometry from IMU and wheel encoder measurements for odometry. We demonstrate in Section \ref{sec:experiments}  that the DLL algorithm performs much better than the 3D  Monte Carlo Localization (MCL) method \cite{Perezgrau17JARS, 10871015} and achieves comparable or better precision to other map-based localization approaches, such as solutions based on the Iterative Closest Point (ICP) algorithm \cite{Rusu_ICRA2011_PCL} or the Normal Distributions Transform  (NDT) \cite{magnusson2007scan}, while running one order of magnitude faster. 

\section{Navigation System}
\label{sec:navigation}

The autonomous navigation system enables the UGV and UAV to perform coordinated navigation in order to safely execute an inspection task. Given the inspection plan as a sequence of PoIs with orientation to be visited, we generate a collision-free trajectory to be tracked for both the UAV and the UGV platforms \revD{by using a modification of the planner described in our previous work \cite{smartinezr2023}. Note that this approach generates trajectories in which all the components of the system (UAV, UGV and tether) do not collide with any obstacles. While some approaches may allow the cable to collide with obstacles \cite{tape_tether_aware}, we have experimentally found that these collisions may lead to unexpected entanglements of the cable with small obstacles (e.g., screws, hooks, etc.). These entanglements are difficult to detect, and planning and executing the required unraveling maneuvers is even more challenging. Therefore, we avoid any contact between the tether and obstacles to ensure safe autonomous navigation of our marsupial system.} 

\revD{The proposed planner includes a global path planner (Section \ref{sec:path_planning}) and a non-linear optimizer (Section \ref{sec:trajectory_planning})} that refines the path into an optimized trajectory. As the \rev{LTS4} system is able to keep the tether taut (see Section \ref{sec:tether_section}), the planner proposed in \cite{smartinezr2023} has been modified to consider that \rev{at any moment} the length of the cable equals the distance between the UGV and the UAV platform and \rev{LTS4}. Therefore, only configurations with a direct Line of Sight (LoS) between these platforms are considered \revD{collision-free}. Finally, the computed trajectory is followed by the trajectory tracker, which is detailed in Section \ref{sec:trajectory_tracking}.

%-------------------------------------------------------------------------------------------------------

\subsection{Trajectory Planning Problem Formulation}

We define a state in the state space as the combination of the position of the UGV 
 $\mathbf{p}_g=(x_g,y_g,z_g)^{T}$, the UAV $\mathbf{p}_a=(x_a,y_a,z_a)^{T}$. At any instant, the tether length $l$ is equal to the distance from the UGV to the UAV ($l = \|\mathbf{p}_a - \mathbf{p}_g\|$). Therefore, in contrast to \cite{smartinezr2023} the length of the tether is not considered as a planner variable but rather considered taut during the whole mission, and thus we have to check for LoS visibility between the platforms to ensure collision-free operation.

Our motion planning algorithm determines the trajectory for the UGV $\mathbf{p}_g(t)$, the UAV $\mathbf{p}_a(t)$ so that the UAV reaches a given goal position, avoiding obstacles. We work on time-discrete trajectories  (\ref{eq:traj_params}), forming a set of states for each time step $i$.

\begin{equation}
\label{eq:traj_params}
    O = \{\mathbf{p}^i_g,\mathbf{p}^i_a,\Delta t^i\}_{i=1,...,n}
\end{equation}

\noindent where $i$ is a time step of the trajectory and $n$ is the total number of time steps. $\Delta t^{i} = t^{i} - t^{i-1}$ is the time increment between steps $i$ and $i-1$. This value is the same for UGV and UAV trajectories. For each $\mathbf{p}^i_a$, $\mathbf{p}^i_{g}$ and $l^i$, there is a taut tether configuration $T^i$. When needed, we discretize it into a set of $m$ positions $\mathbf{p}_{t}=(x_{t}, y_{t}, z_{t})$ (\ref{eq:tether}).

\begin{equation}
\label{eq:tether}
    T^i = \{\mathbf{p}^j_t\}_{j=1,...,m}
\end{equation}

%-------------------------------------------------------------------------------------------------------

\subsection{Path Planning}
\label{sec:path_planning}

In the first step, the path planner is based on an RRT* algorithm, following the general structure presented in \cite{smartinezr2023} with modifications to accommodate the characteristics of the tethered marsupial system. As we are working with a taut tether, we adapted the algorithms responsible for checking catenary collisions to permit only tether states that ensure LoS between the UGV and UAV platforms, meaning that hanging tether states are not allowed.

We employ the RRT* algorithm \cite{karaman_rrt_star} to solve the problem due to its ability to deal with high-dimensional spaces. The RRT* generates a path in a six-dimensional space composed of the UGV and UAV positions $\mathbf{x}=\{\mathbf{p}_g,\mathbf{p}_a\}$, from an initial state $\mathbf{x}^i$ to a goal state $\mathbf{x}^g$ in which only the aerial robot position is set (the final position of the UGV will depend on the UAV goal and the environment). The tether is just evaluated procedurally in the RRT* \texttt{Steering} stage, through the \texttt{checkTetherFeasibility} algorithm, presented in Section~\ref{sec:checktether}. The RRT* cost function is just a weighted sum of the total length of the UAV and UGV paths.

The planner leverages a 3D point cloud to represent the environment for UAV and UGV path planning. Initially, a traversability analysis, similar to the one carried out in~\cite{driving_pc}, identifies traversable areas for the UGV, which are used to generate new positions for the UGV. For the UAV, \revD{we compute the EDF from the full 3D point cloud, obtaining a grid representation that stores the distances from the grid position to its closest obstacle. New samples of the position of the UAV are obtained randomly on the grid points in which the EDF exceeds a safety distance.}

\begin{figure}[H]
 %   \centering
    \includegraphics[width=1.0\linewidth]{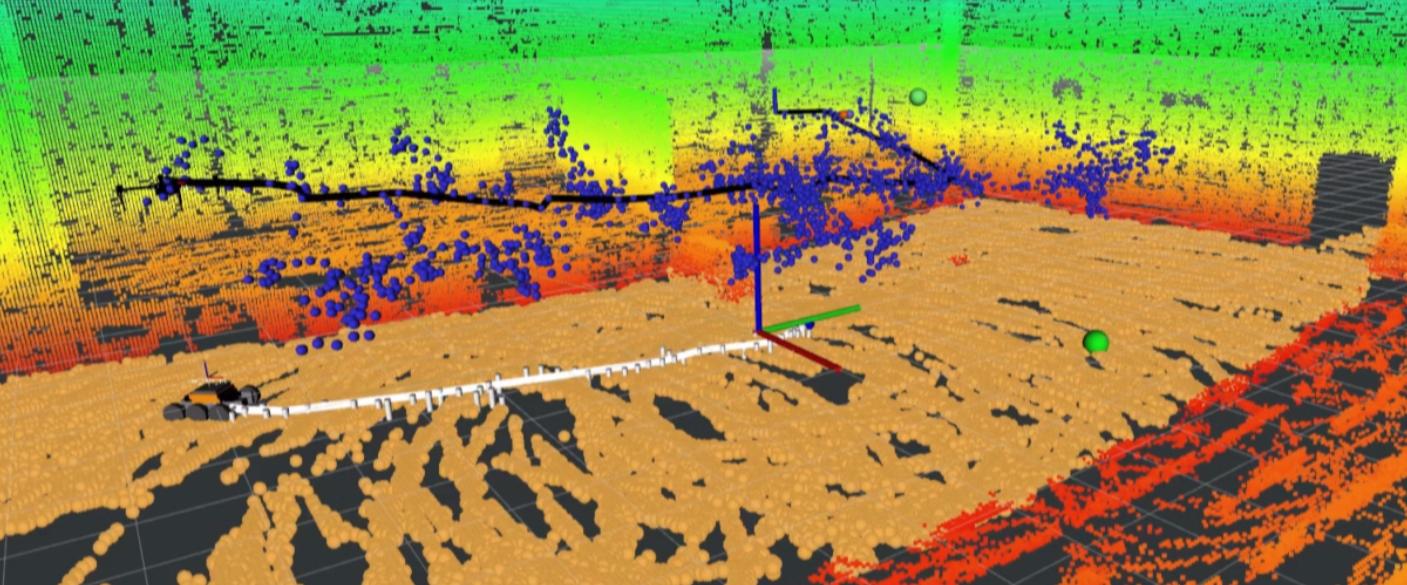}
    \caption{Example of the path obtained by the proposed path planner based on the RRT* algorithm in the Theater scenario. Points in blue and white color represent explored nodes, i.e., 
 nodes of the RRT*'s tree, for the UAV and the UGV, respectively. Black and white lines represent the obtained path for the UAV and the UGV, respectively. The green point represents the random point generated in the current iteration.} 
\label{fig:planner}
\end{figure}

Figure \ref{fig:planner} shows an example of a path trajectory generated by the path planner. The planner generates a sequence of waypoints for the UAV (represented with a dotted black line) and the UGV (dotted white line) with direct visibility between the UAV and the UGV at each corresponding waypoint, checked in the \texttt{checkTetherFeasibility}  algorithm.

\subsubsection{checkTetherFeasibility Algorithm}
\label{sec:checktether}

This algorithm is integrated into the RRT* steering process and verifies if a collision-free tether exists between the UGV and UAV. For each new state, the tether length $l^i$ is defined as the Euclidean distance between $\mathbf{p}^i_a$ and $\mathbf{p}^i_g$. If the tether is feasible, the algorithm marks the corresponding UGV and UAV positions as valid for $l^i$, returning \texttt{true}.

\subsection{Trajectory Planning}
\label{sec:trajectory_planning}

The trajectory planning improves the initial path obtained by the path planner (see previous section) by solving a non-linear optimization problem that considers the time dimension as well as safety and geometrical constraints. The path planning method from the previous section outputs a sequence of collision-free robot positions $\{\mathbf{p}^i_g,\mathbf{p}^i_a\}_{i=1,...,n}$, where $n$ is the trajectory length. This sequence, however, does not include time-related information as in (\ref{eq:traj_params}), nor does it consider time or safety constraints.

The method presented in \cite{smartinezr2023} incorporates time information into the initial path to create an initial trajectory solution. To achieve this, the method spaces the waypoints equidistantly along the path. Then, the $\Delta t^{i}$ values are initialized by assigning constant scalar speeds, $v_g$ and $v_a$, to the UGV and UAV paths, respectively. Both trajectories use the same $\Delta t^{i}$ to ensure that the UGV and UAV reach each waypoint simultaneously. Thus, the value of $\Delta t^{i}$  for each state in (\ref{eq:traj_params}) is the largest value between  $\|\mathbf{p}^i_g - \mathbf{p}^{i+1}_g\|/v_g$ and $\|\mathbf{p}^i_a - \mathbf{p}^{i+1}_a\|/v_a$. 

As the trajectory is computed through optimization, we retain most constraints from our method proposed in \cite{smartinezr2023}. These include limits on velocity and acceleration, equidistant spacing between consecutive robot poses, maximization of the distance from obstacles for each agent (UGV, UAV, and tether), minimization of trajectory execution time, and maximization of smoothness. 

\begin{figure}[H]
  %  \centering
    \includegraphics[width=1.0\linewidth]{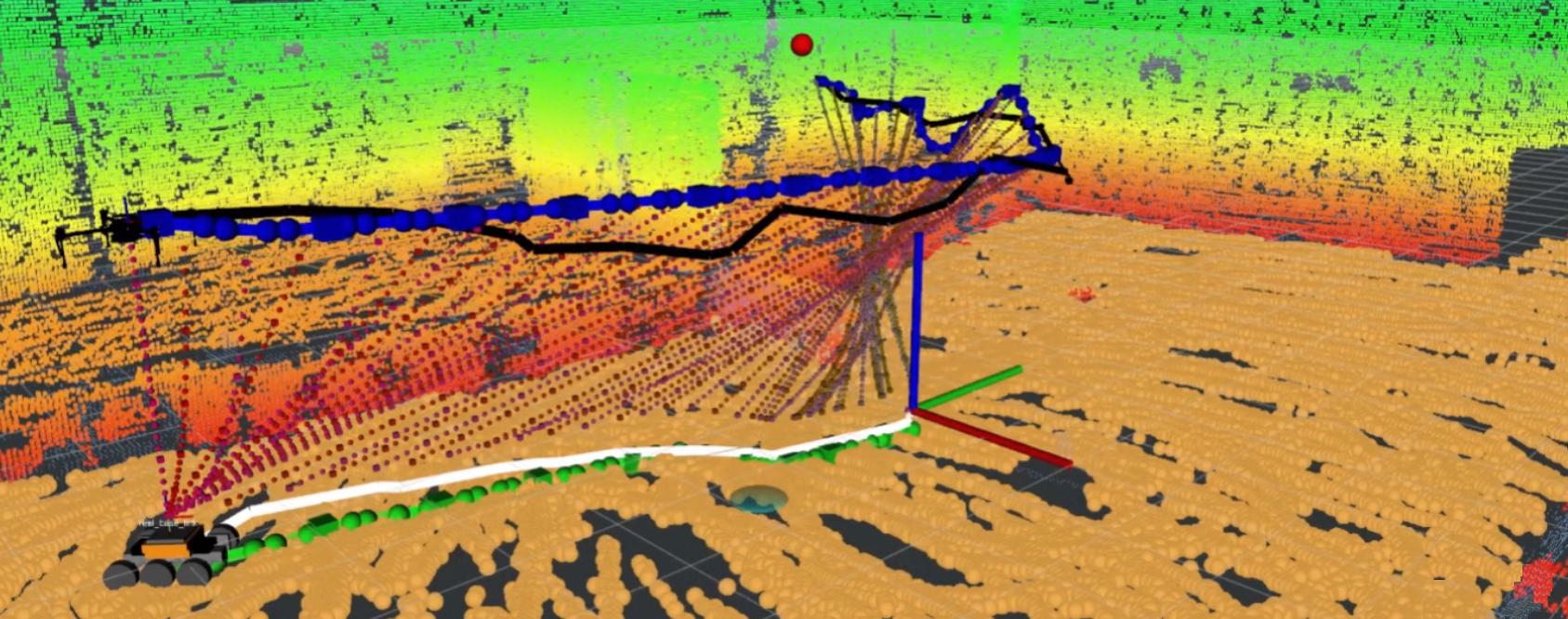}
    \caption{Example of final trajectory computed by our non-linear optimizer. The black line and blue dots represent the path and the trajectory, respectively, for the UAV. The white line and green dots represent the path and the trajectory, respectively, for the UGV. Red dots represent the tether of the optimized~trajectory.} 
\label{fig:optimizer}
\end{figure}

Finally, the optimizer outputs a safer (more distant from obstacles) and smoother trajectory than the initial one, along with speeds and accelerations for traversal, as shown in Figure \ref{fig:optimizer}.

In the particular implementation used in this paper, the primary differences from the base method \cite{smartinezr2023} lie in the tether-related constraints. First, the Unfeasible Tether Length Constraint penalizes tether lengths that exceed the maximum feasible length. Second, for the Tether Obstacle Avoidance Constraint, we evaluate only the collision-free LoS between platforms, using a straight-line approximation rather than calculating the catenary. The details of these constraints are presented in Sections \ref{sec:tether_length} and \ref{sec:tether_obstacle}, respectively.

\subsubsection{Unfeasible Tether Length Constraint}
\label{sec:tether_length}
 
Equation (\ref{eq:eq_length}) penalizes the distances $d^i_u$ between $\mathbf{p}^i_g$  and $\mathbf{p}^i_a$ are longer than the maximum feasible length $L_{max}$. 

\begin{eqnarray}
  \label{eq:eq_length}
  \delta^i_u =& \left \{
  \begin{array}{cc}
  e^{d^i_u - L_{max} } -1 & ,\textrm{if}\  \  d^i_u \ > \ L_{max} \ \\
  0 &, \textrm{otherwise}
  \end{array}
  \right . 
\end{eqnarray}

%---------------------------------------------------
\subsubsection{Tether Obstacle Avoidance Constraint}
\label{sec:tether_obstacle}
To check for collisions on the tether, we collect $m$ samples of the tether as expressed in (\ref{eq:tether}). Assuming a taut tether, we have to sample the straight line $l^{i}$ connecting $\mathbf{p}^i_g$ to $\mathbf{p}^i_a$. Then, we obtain the distance to the nearest obstacle of each sample from the EDF, $d^i_{ot,j}$. Finally, the residual expressed in (\ref{eq:eq_tether_obst}) is the sum of the inverse of the distances. We increase the weight of those samples closer than a safety distance $\rho_{ot}$ to guarantee higher costs in these cases using $\rho_{j}= \beta$, with $\beta >> 1$.

\begin{eqnarray}
  \label{eq:eq_tether_obst}
  \delta^i_{ot} &=& \sum_{j=1}^{m} \frac{\rho_{j}}{d^i_{ot,j}}  , \  \rho_{j} = \left \{
      \begin{array}{cc}
      1 & ,\mathrm{if}\  \  d^i_{ot,j}  >  \rho_{ot}\ \\
      \beta & , \textrm{otherwise}
  \end{array}
  \right .
\end{eqnarray}

%-------------------------------------------------------------------------------------------------------
\subsection{Trajectory Tracking}
\label{sec:trajectory_tracking}

In this section, we detail the control system designed to track the trajectories generated in Section \ref{sec:trajectory_planning}. We use a distributed approach in which each platform has its independent trajectory tracker. Then, a coordination module ensures the synchronization of their~movements.

\begin{figure}[H]
   % \centering
    \includegraphics[width=0.8\linewidth]{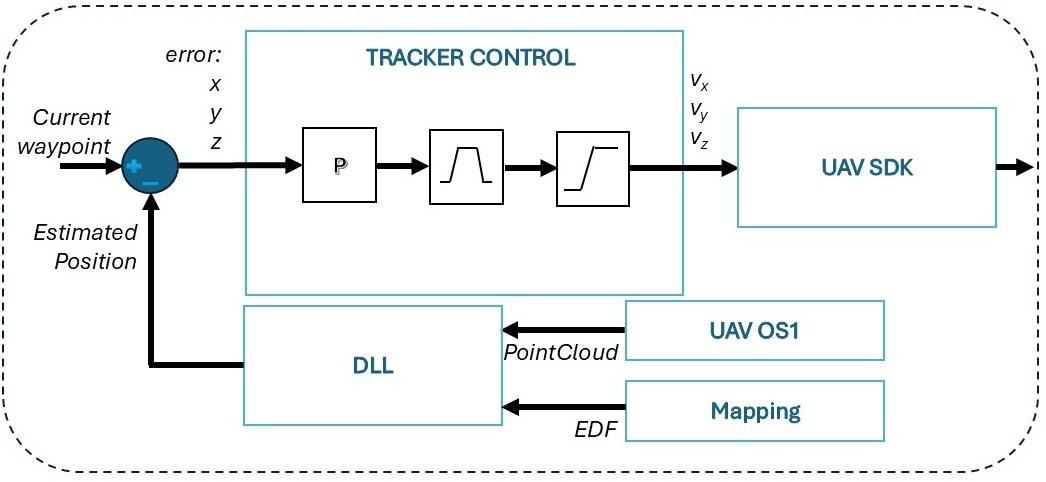}
 %   \centering
    \caption{System diagram of the proposed path tracker.} 
\label{fig:tracker}
\end{figure}

Thus, for trajectory tracking, we assume that both the UAV and the UGV accept holonomic velocity commands. We have implemented a simple trajectory tracking scheme based on proportional control of each one of the coordinate axes (see Figure \ref{fig:tracker}). Our trajectory tracker acts similarly in both UAV and UGV platforms but adds a proportional height controller in the case of the UAV. Finally, we incorporate an additional yaw control in the UAV, which is used when reaching the inspection point to ensure that the vehicle is pointing towards the area of interest.  

Regarding the magnitude of the velocity command vector ($\boldsymbol v$) in the lateral motion, we make use of trapezoidal velocity profiles for waypoint tracking. The maximum velocity for the trapezoid is set to the planned $\rho_{vg}$ and $\rho_{va}$, respectively, but the actual mean velocity is slightly less than the commanded due to the velocity profile.

The maximum commanded speed has been limited to match the speed of the slowest subsystem, which in this implementation is the UGV. This decision was made for safety reasons and to facilitate the coordination of the vehicles. In the experiments, we set the speeds of both the UGV and the UAV ($\rho_{vg}$ and $\rho_{va}$, respectively,) to 0.25 m/s.

Finally, in order to track the trajectory computed by our approach, the UGV and the UAV should follow the trajectories coordinately. This task is carried out by our TCM (see Figure \ref{fig:software}). We opt for a loosely coupled solution based on time synchronization. TCM can only command new waypoints to both the UGV and UAV simultaneously, provided that both platforms have successfully reached the previous one. Thus, when a subsystem reaches its current waypoint, it waits for the other in case of need.

%%%%%%%%%%%%%%%%%%%%%%%%%%%%%%%%%%%%%%%%%%
\section{Experimental Results}
\label{sec:experiments}

We carried out different field experiments with our marsupial configuration in three different buildings of the UPO, Seville (Spain). Scenario 1 is situated in Building 45, home to the Service Robotics Laboratory facilities. Next, we conducted additional tests in two old, abandoned buildings on the same university campus, both of which show significant structural damage. Scenario 2 takes place within the Abandoned Thermal Station, whereas Scenario 3 is set in the Old Theater building. The logs of all experiments are accessible through our institutional repository (\url{https://robotics.upo.es/datasets/marsupial}, accessed on 3 Nov 2025). Furthermore, additional experimental details are presented in the video accompanying this paper.

\subsection{Scenario 1. Experiments 1--3: Flight Duration Tests}
\label{sec:scenario1}
\begin{figure}[H]
    %\centering
   
\centering %% If there is a figure in wide page, please release command \centering
 \begin{tabular}{cc}
    
    \includegraphics[width=0.485\textwidth]{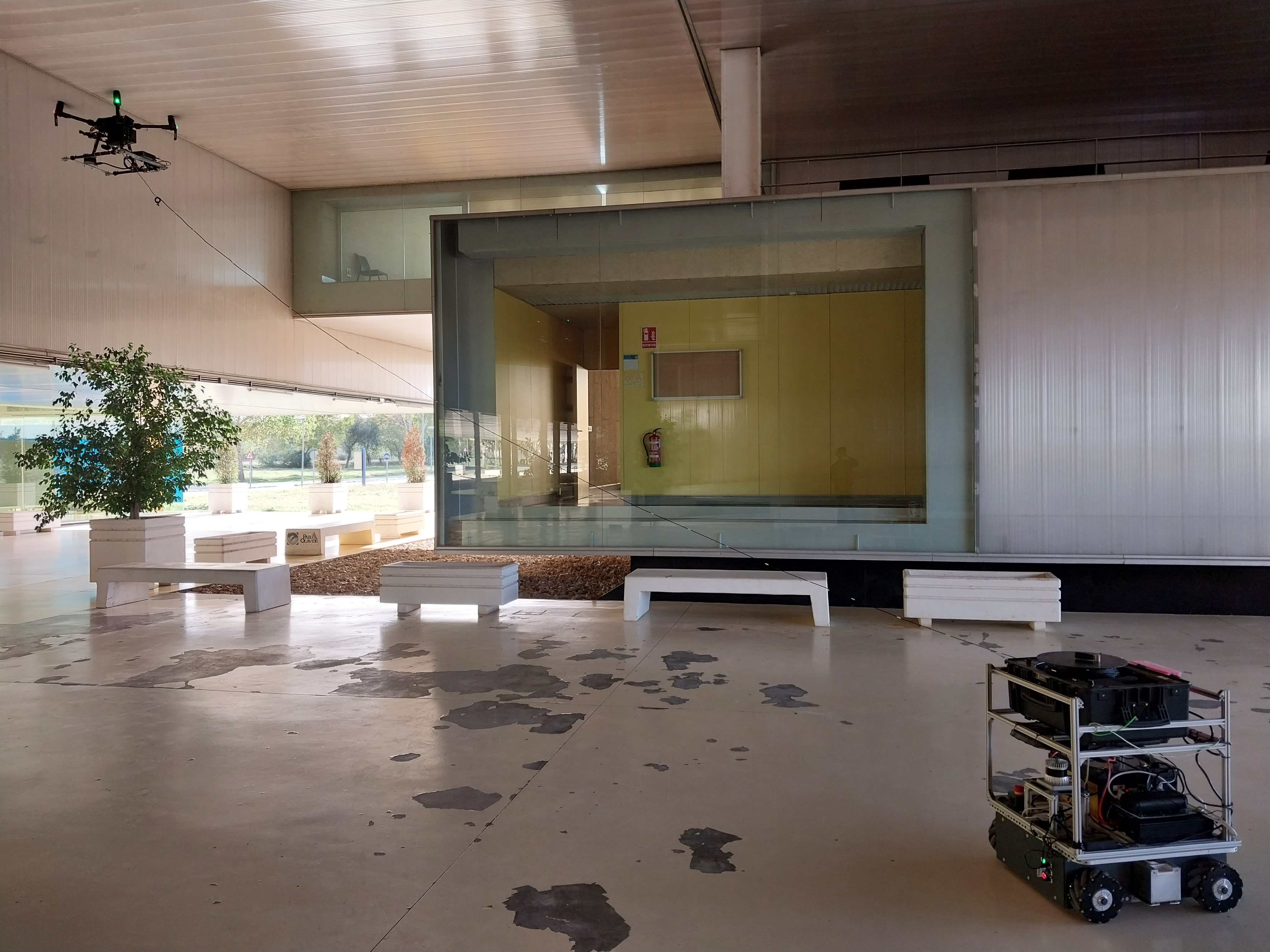} &
    \includegraphics[width=0.485\textwidth]{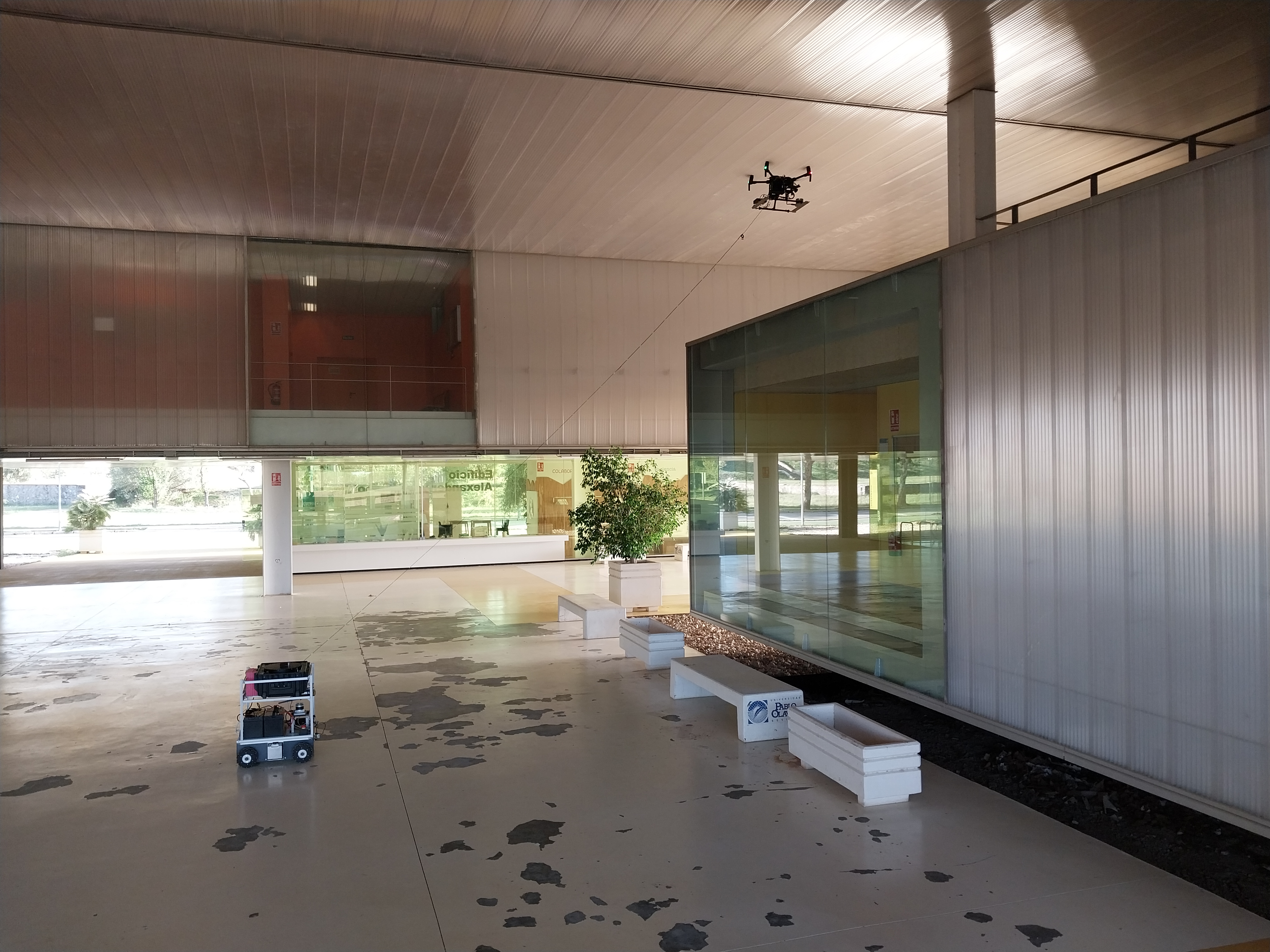}

    \end{tabular}
    \caption{Two snapshots of one flight duration test at building 45.}
    \label{fig:experiment1}
\end{figure}

\revD{The main goals of this experiment were to perform the necessary system integrity checks of the whole marsupial system and to estimate the flight endurance of the system. We performed three different experiments (Experiments 1--3) in which our marsupial system was commanded in manual mode, mimicking the stop-and-go behavior in an inspection mission. The duration of all the experiments exceeded one hour, with Experiment 3 exceeding two hours. All the experiments were performed in the same load condition. Figure \ref{fig:experiment1} shows two snapshots obtained during the experiments.}

\begin{figure}[H]
   % \centering
    \includegraphics[width=0.7\linewidth]{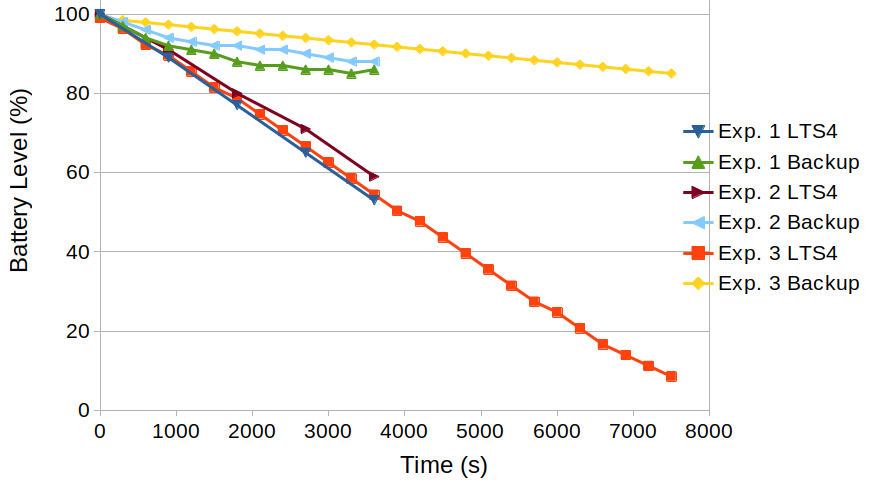}
    \caption{Remaining battery levels of LTS4 and backup batteries in Experiments 1--3.}
    \label{fig:plot_duration}
\end{figure}

\revD{Figure \ref{fig:plot_duration} represents the level of the batteries powering the LTS4 and the onboard UAV backup battery in Experiments 1--3. Remarkably, the discharge rate of the LTS4 battery remained approximately constant throughout all three experiments; the discharge is 46\% in one hour in the worst case. Regarding the backup battery, the remaining level was above 80\% in all three cases, which indicates that there was not significant usage, as the LTS4 system worked without issues during all three experiments. Moreover, this remaining battery level would allow the monitoring operator to safely land the UAV in the event of an LTS4 malfunction, as was designed in Section \ref{sec:power}. With all the gathered data, we can conclude that our proposed system can perform inspection missions lasting up to two~hours.}

\subsection{Scenario 2: Abandoned Thermal Station}
\label{sec:scenario2}
Scenario 2 is located at the Abandoned Thermal Station building of the UPO, built in the decade of the 1950s and \revD{currently  exhibiting significant structural damage and surface debris. These conditions} prevented our ARCO platform \revD{from autonomously navigating within the building. Consequently, we designed a set of experiments focused on the UAV system to evaluate its indoor localization and navigation capabilities.}

\subsubsection{Experiments 4--5: UAV Localization Tests}

\revD{In Experiments 4 and 5,} \rev{we tested the precision of our localization system} in manual mode. To this end, we designed a challenging experiment in which we made our UAV move around the whole experimental scenario at different height levels while recording the data from the sensors. Then, we tested different localization methods to compare their results and their average computation time. To evaluate the precision of the localization algorithm, we obtained the distance from each LiDAR detection inside a measurement to the closest obstacle in the map to check for errors in the localization. The lower the average error in a measurement is, the better the vehicle is localized in the environment. This was necessary due to the lack of ground truth from external systems available. Table \ref{tab:localization} represents the mean distance from the LiDAR points to the closest obstacle with different localization methods and their average execution time. We configured the MCL method with 500 particles, and we considered that each measure of the OS1 sensor has a standard deviation of 0.1 m and that the odometry has a relative error of 10\%. We used the ICP implementation of the Point Cloud Library (PCL) with a maximum of 50 iterations, a maximum correspondence distance of 0.1 m and a RANSAC outlier rejection threshold of one meter. Finally, our DLL solver was implemented with the Ceres library, being configured to use 8 threads. \revD{All the experiments were executed in the NUC11TNKi5 computer onboard the UAV, which has a 13th Gen Intel\textregistered\  Core\texttrademark \ processor with 12 cores, one thread per core, and 16 GB of RAM.}

Results indicate that our DLL method outperforms the MCL in terms of precision and runs significantly faster when compared to the ICP method, achieving similar errors. 
\revD{While these results offer an indirect evaluation of the accuracy of the localization, in practice we found that it was enough to safely perform cooperative missions in our marsupial system, making 
DLL the better method for real-time localization.}

For more details about the experiment, please refer to the attached video, where you can check the behavior of our DLL in the localization tests.

\begin{table}[t!]
   \caption{Average computation time of the update step on each method and average distance from each point of the point-cloud to the closest obstacle in the map in Scenario 2.}
   \label{tab:localization}
   \centering
   \begin{tabular}{|c|c|c|c|}

   \hline Exp. & Method & Comp. time (s) & Avg. Dist. (m) \\ \hline

   &MCL-3D & \revD{0.71}  & 0.45 \\
   4&ICP & \revD{1.8} & 0.13 \\
   &DLL & \revD{0.07} & 0.14 \\ \hline
   &\revD{MCL-3D} & \revD{0.64}  & \revD{1.25} \\
   5&\revD{ICP} & \revD{2.9} & \revD{0.56} \\
   &\revD{DLL} & \revD{0.11} & \revD{0.26} \\
   \hline

   \end{tabular}
\end{table}

\subsubsection{Experiment 6: UAV Autonomous Flight Test}

\begin{figure}[H]
    \includegraphics[width=0.9\textwidth]{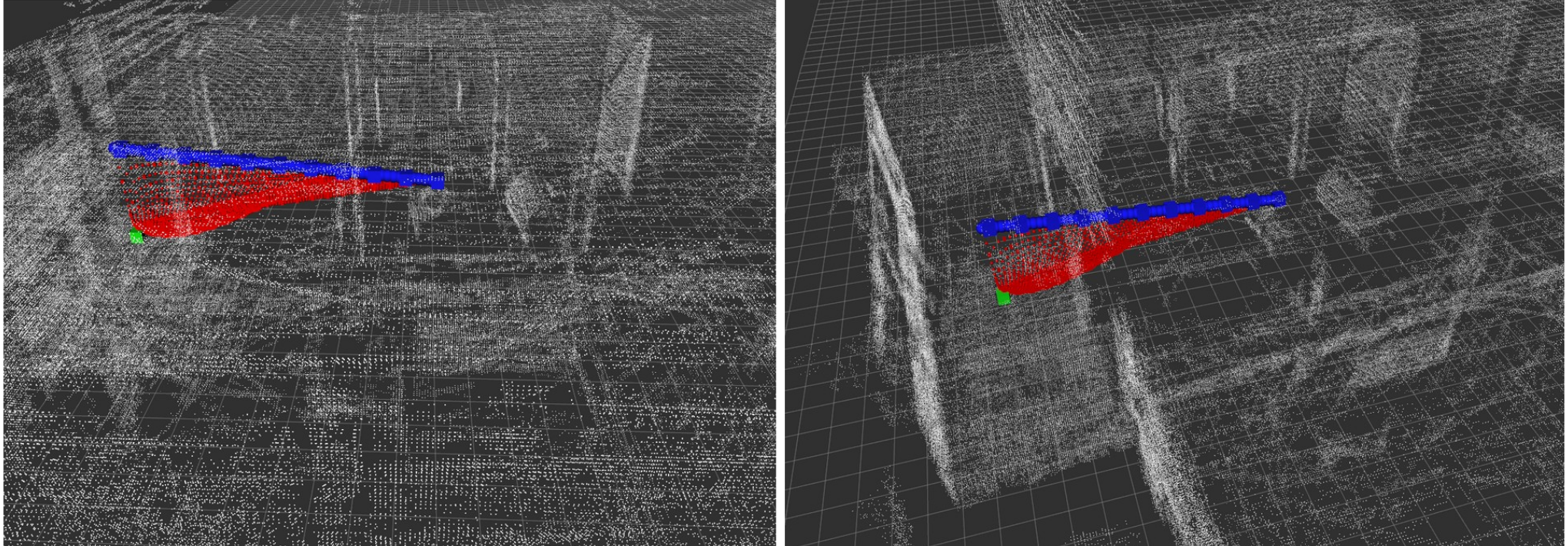}
    \caption{Reference trajectory (blue line) tracking experiments in Scenario 2. The position of the reel system is on a green dot, and the red lines represent the tether configurations during the \rev{autonomous flight test}. The points in white represent the 3D map of Scenario 2.}
    \label{fig:uav_trajectory}
\end{figure}

\revD{After 
 demonstrating that our localization system could accurately estimate the robot’s pose in real time, we conducted Experiment 6 to evaluate the UAV’s trajectory tracking system. To this end, a simple inspection plan was designed, and the trajectory to be tracked by the UAV was generated using our trajectory planner, assuming a static UGV}. Figure \ref{fig:uav_trajectory} represents the generated trajectory to perform the UAV autonomous navigation test represented in the 3D model of the environment. The tracking system was able to autonomously follow the trajectory in all the executions, \revD{achieving mean lateral cross-track and altitude} errors of 0.07 m and 0.08 m, respectively.

Please refer to the attached video for \revD{additional details regarding} the performance of the trajectory tracker.

\subsection{Scenario 3. Experiment 7: Emulated Inspection Test}
\label{sec:scenario3}

Finally, to test the whole system in autonomous mode, we emulated a structural assessment inspection \revD{in Experiment 7, which was carried out in Scenario 3. In Experiment 7}, our marsupial system is \revD{commanded to execute} a mission designed to look for emulated defects in the Old Theater building of the UPO. To emulate the defects, we \revD{deployed} an array of twelve Augmented Reality (AR) markers with a side of fifteen centimeters on one of the walls of the building (see Figure \ref{fig:system_proposed}b). We equipped the UAV with a high-resolution camera to automatically detect the markers by using the ArUco library \cite{aruco}.

\begin{figure}[H]
  %  \centering
    \includegraphics[width=\linewidth]{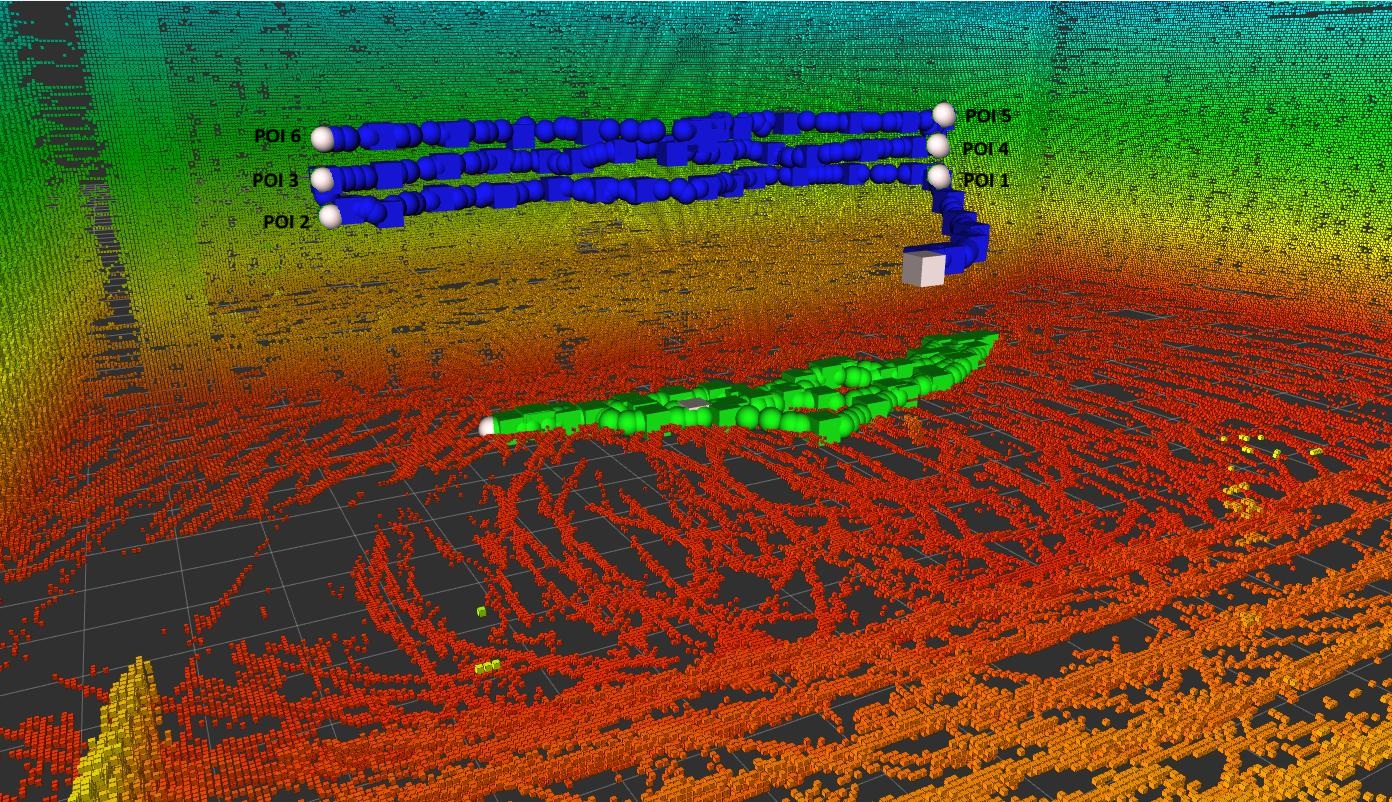}
    \caption{Computed trajectory  for the inspection mission of UGV (in green dots) and UAV (in blue dots). The square in white represents the initial UAV position. The spheres in white represent the PoIs to be visited.}
    \label{fig:trajectory_wall}
\end{figure}

Figure \ref{fig:maps_example}a represents an obtained 3D map of the environment, which was used for localization and path planning purposes on both platforms in Scenario 3. The goal of the experiment is to make an in-depth inspection of one of the walls of the building. To this end, we \revD{distributed} the PoIs of our inspection plan following a zig-zag line. \revD{The distribution of the PoIs was sufficient to ensure complete wall coverage with a spatial resolution of twenty pixels per square centimeter, which is adequate for the detection of the emulated defects. We used the Fields2Cover library \cite{Mier_Fields2Cover_An_open-source_2023} to generate the sequence of PoIs (see Figure \ref{fig:trajectory_wall}).}

\subsubsection{Trajectory Planning Results}

With the inspection plan defined as a sequence of PoIs, we generated a trajectory for the marsupial system using our trajectory planner module (see Section \ref{sec:trajectory_planning}). Note that we need multiple calls to the planner to generate trajectories connecting the different PoIs.

Table \ref{tab:planning} presents the statistics of our trajectory planning method compared to the approach proposed in \cite{smartinezr2023}, which accounts for a non-straight tether when generating the complete inspection mission. Both methods successfully generated trajectories. Notably, our planning method achieves lower execution times compared to the baseline.

Regarding clearance, our method yields a smaller mean distance between the tether and both the obstacles and the UAV. This is expected, as we consider only a straight tether, which reduces the available maneuvering space. Finally, in terms of total traveled distance, both methods produce similar UAV paths; however, our method requires the UGV to move slightly more due to the LoS constraint.

In conclusion, our method is capable of generating safe trajectories for the system with lower computation time compared to the approach in \cite{smartinezr2023}.

\begin{table}[t!]

\caption{Trajectory planning statistics for the proposed planner (LoS) and the planner proposed in \cite{smartinezr2023} (catenary).}
\label{tab:planning}
\begin{tabular}{|c|c|ccc|cc|} \hline
   Method & Execution time (s) & \multicolumn{3}{c|}{Avg. distance to obstacles (m)} & \multicolumn{2}{c|}{Trajectory Length (m)} \\  & & Cable & UAV & UGV & UAV & UGV \\\hline
    Catenary & 3.37 & 1.51 & 1.54 & 1.09 & 50.40 & 11.59   \\ \hline
   LoS & 0.49 & 1.11 & 1.08 & 1.65 & 48.30 & 14.14 \\
   
   \hline
\end{tabular}
\end{table}

\subsubsection{Mission Execution Results}

\begin{figure}[H]
    \includegraphics[width=0.7\linewidth]{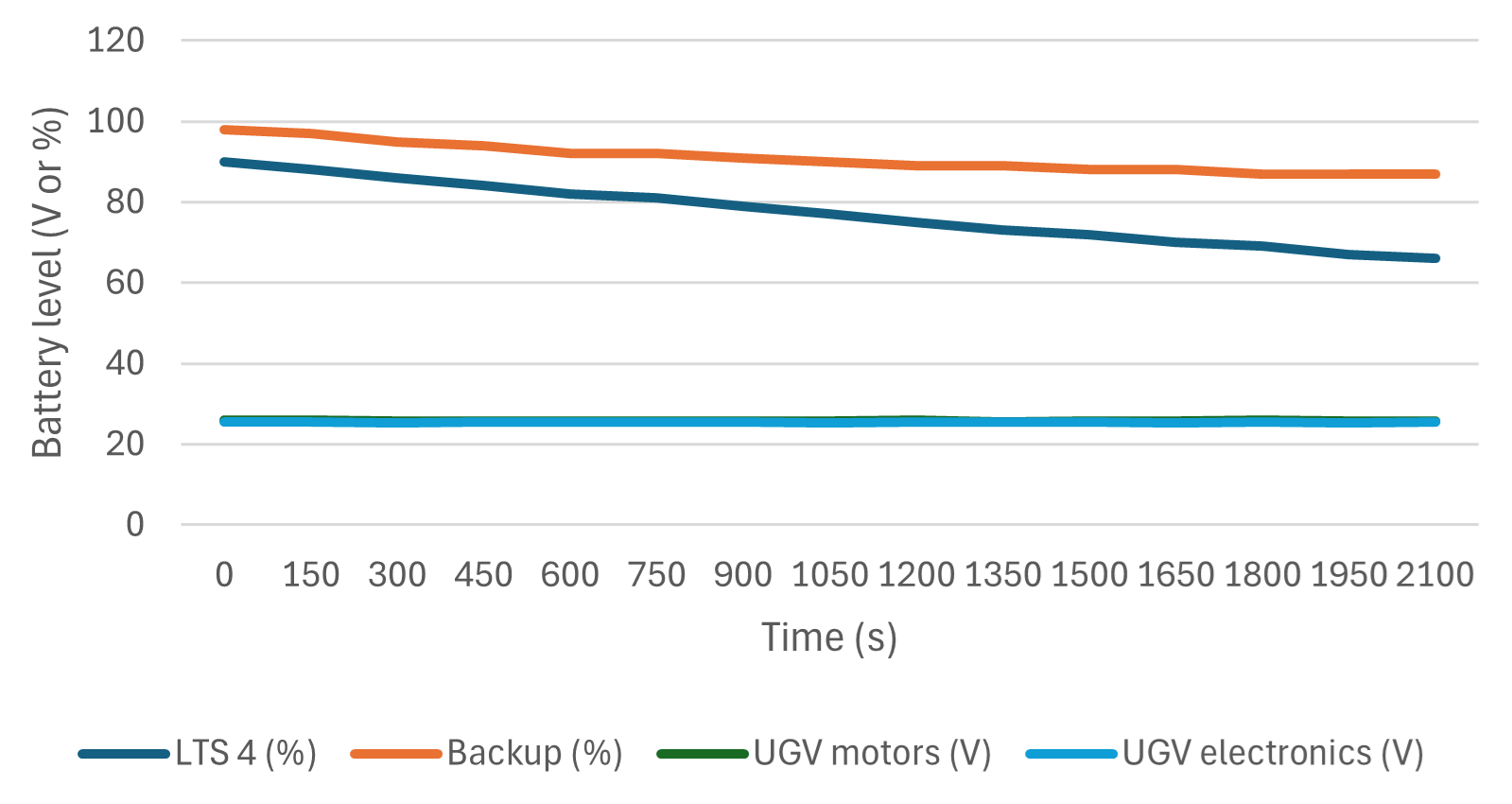}
    \caption{Remaining battery levels of the LTS4 (red and blue) and backup batteries and measured voltage of the UGV batteries (green and cyan) in Experiment 7.}
    \label{fig:discharge_experiment}
\end{figure}

In the experimental session, our battery-powered LTS4 system allowed us to execute the inspection mission three times without the need of changing any batteries, for a total duration of thirty-five minutes. The trajectories commanded to the UAV and the UGV are represented in Figure \ref{fig:trajectory_wall} and are composed of 245 waypoints. As detailed in Section~\ref{sec:power}, a safety operator supervised the execution of the experiment, ensuring the absence of unexpected tether entanglements and monitoring the charge level of the UAV’s onboard backup battery. Figure \ref{fig:discharge_experiment} illustrates the battery levels of the UAV, the UGV, and the batteries supplying the LTS4 system. The backup battery level remained consistently above 80\% throughout the experiment. Moreover, since no unexpected entanglements were observed, no operator intervention was required. Regarding the UGV’s onboard batteries, their voltage level remained approximately constant at 25.5 V during the entire experiment. Although voltage is not a precise indicator of the remaining charge for LiFePO4 batteries, this stability suggests that the battery operated within its mid-range (approximately 30 to 70\%). Finally, the batteries powering the LTS4 maintained a charge level above 60\%, indicating that safe operation could have continued for more than one hour after the~experiment.  

The performance of our trajectory tracking systems was also remarkable. The system was capable of reaching all the planned waypoints in all three experimental runs. Finally, we evaluate the performance of the TCM. Table \ref{tab:waiting} presents the statistics of the waiting times for both the UAV and the UGV during each mission execution. The results are highly consistent. The UAV experiences a higher number of waiting events due to its lower velocity. However, the difference in waiting times is small, averaging approximately 1.5 s. Although this value may seem relatively high, it should be noted that the trajectory includes segments where only one vehicle is moving, causing the other to wait until the next waypoint is reached. Finally, the relatively high standard deviation indicates the presence of outliers in which the waiting times are prolonged for several seconds. This typically occurs at the beginning of the experiment.

\begin{table}
   \caption{Average waiting time and standard deviation for the UAV and UGV induced by the TCM on Experiment 7. Values in parentheses indicate the number of waiting events. }
   \label{tab:waiting}
   \begin{tabular}{|c|c|c|} \hline
        Mission execution &UAV waiting time (s) & UGV waiting time (s)  \\ \hline
        1 & 1.89 $\pm$ 2.21 (147) & 1.51 $\pm$ 1.93 (147) \\
        2 & 1.43 $\pm$ 2.73 (144) & 1.43 $\pm$ 2.14 (101) \\
        3 & 1.47 $\pm$ 2.39 (142) & 1.39 $\pm$ 1.78 (103) \\ \hline    
        
   \end{tabular}

\end{table}

\begin{figure}[H]
    \centering
    \includegraphics[width=\linewidth]{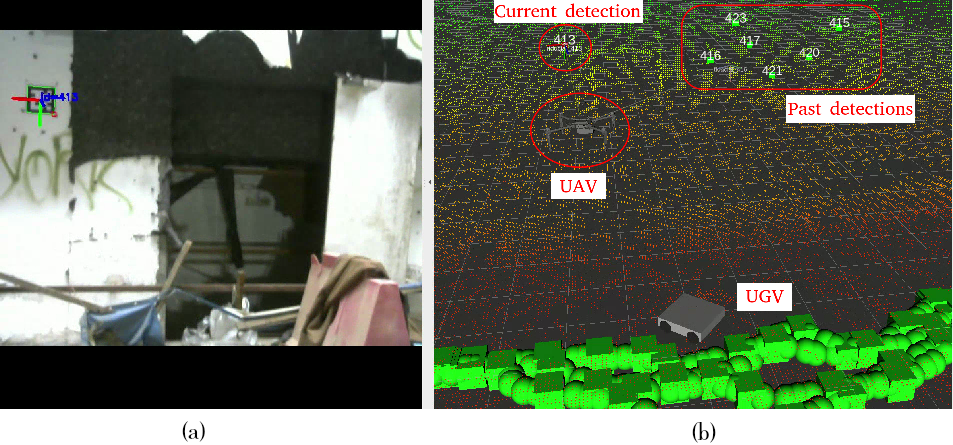}
    \caption{Current  and past detections of emulated defects (AR Markers) obtained in Scenario 3, Experiment 7. (a) Current image from the onboard camera with an overlay of the current detection (ID 413). (b) Estimated positions of the UAV, the UGV and the current and past detections over the 3D map of the environment.}
    \label{fig:detections_aruco}
\end{figure}

The marsupial system was able to detect all the emulated defects in the three experimental runs, as illustrated in Figure \ref{fig:detections_aruco}. Statistics on the error in the estimated position of the detected ArUco markers are provided in Table \ref{tab:aruco}. In particular, the mean detection error of each marker is on the order of centimeters. Moreover, the cases with greater errors are usually related to a low number of observations of the AR marker. \revD{The achieved precision of the estimated position of the AR markers also validates the precision of our localization~system.}

\begin{table}[t!]

\caption{Average error and number of observations in the position estimation of the AR markers.}
\label{tab:aruco}
\begin{tabular}{|c|cccccccccccc|} \hline
   AR ID's & 411 & 412 & 413 & 415 & 416 & 417 & 418 & 420 & 421 & 422 & 423 & 477\\ \hline
   Error (cm) & 4.2 & 12.6 & 13.6 & 19.8 & 9.8 & 6.2 & 8.7 & 11.6 & 18.7 & 3.1 & 13.0  & 7.6 
   \\ \hline
   Observations & 88 & 28 & 277 & 28 & 215 & 445 & 198 & 568 & 344 & 307 & 205 & 29 \\
   
   \hline
\end{tabular}

\end{table}

%%%%%%%%%%%%%%%%%%%%%%%%%%%%%%%%%%%%%%%%%%
\section{Discussion}

\label{sec:discussion}

This paper has presented the design, implementation, and validation of a marsupial robotic system capable of performing long-duration inspections in GNSS-denied environments. Experimental results confirm the hypothesis that a tethered UAV-UGV system \revD{can extend UAV flight time to more than two hours of continuous operation}, opening the door to long-duration inspections. The accuracy in detecting AR markers, with a mean error in the order of centimeters, demonstrates the viability of the system for detailed inspection~tasks.

\revD{%Even though power consumption was higher than expected, 
The platform proved capable of performing inspection missions lasting more than two hours, which aligns with our estimates when accounting for the power consumption of the UGV’s internal LiDAR and router, both of which are connected to the inverter.} However, the flight duration \revD{could} be further extended by \revD{integrating} more battery packs to the UGV, \revD{as the platform remains well below its payload limit}.

\revD{With the knowledge gathered from our experimentation, we are confident that the proposed marsupial system represents a viable alternative with several advantages over existing approaches in the literature. One of its key strengths lies in the capability of the UGV to supply power to the UAV through the tether, which significantly extends the UAV’s endurance beyond the limitations imposed by its onboard battery. In practice, this enables flight times of over two hours, a remarkable improvement over conventional untethered UAVs. Furthermore, in contrast to systems where the UAV is tethered to a fixed station, our mobile UGV base provides additional flexibility by enabling the UAV to cover a broader range of inspection areas. The complementary nature of both platforms enhances mission capabilities: while the UAV contributes aerial mobility and monitoring functionality, the UGV acts as a stable power source and offers additional payload capacity. Marsupial UAV–UGV systems have already demonstrated their utility in hazardous and confined environments, as shown, for instance, by the CSIRO team in the DARPA Subterranean Challenge \cite{CSIRO2024} and the Oxpecter platform \cite{oxpecter_drones7020073}. Our approach significantly advances autonomy,  demonstrating long-endurance inspection in GNSS-denied environments with precise localization and safe trajectory planning.}

Despite these benefits, our approach is not without limitations.  While it is true that tethering limits the UAV’s maneuverability and may constrain its flyable space, requiring careful management to prevent collisions, the extended flight duration enables our solution to carry out long-duration inspection missions, where steady flight is essential to ensure the validity of the measurements. In our view, the extended flight time outweighs the reduction in maneuverability and speed for the applications considered. Maintaining proper tension is essential, since downward tension directly affects payload capacity, whereas lateral forces demand corrective control actions from the UAV to maintain stability. However, we found that by configuring LTS4 to perform a tension of 1 N to the UAV, we can keep the cable straight when only increasing the required force to hover in a 2.
5\%, as the UAV weighs approximately four kilograms.

\revD{Another drawback is that the tether carries the risk of becoming entangled or snagged on environmental obstacles. To mitigate this, missions are carefully designed to prevent any contact between the tether and the environment. The cable is continuously monitored during the experiments, which are halted if entanglements or other unexpected events occur. Finally, while our system achieves effective cooperation between the UAV and UGV, further improvements could be obtained through variable-speed trajectory tracking, which would enable smoother coordination between both platforms.}

%%%%%%%%%%%%%%%%%%%%%%%%%%%%%%%%%%%%%%%%%%
\section{Conclusions}
\label{sec:conclusions}

This paper presents our solution for long-duration inspection: a tethered marsupial system connecting a UAV \revD{to} a UGV. In it, we detail the hardware and software configurations that allow autonomous navigation in UAV inspection tasks, as demonstrated in several field experiments. Notably, \revD{we have demonstrated that the proposed system has a flight duration of over two hours} thanks to the power supplied by the batteries onboard the UGV. Therefore, the proposed marsupial system successfully extends the operational capabilities of the UAV.

The \revD{proposed} tethered marsupial system can successfully perform autonomous inspection tasks, detecting targets such as markers without being hindered by the tether. The inspection tasks are carried out with high accuracy both in terms of the localization of the marsupial robot and in the detection of the markers.

\subsection*{Future Work}

Future work will focus on refining the power system to further extend the operational range. Additionally, we are developing a continuous control system within the TCM to enable smooth coordination between platforms by adjusting the tracking velocity of each system, moving away from the current stop-and-go synchronization.

\revD{Moreover, we aim to expand the inspection capabilities of the system by integrating additional sensors required for advanced inspection tasks}, such as radar and IR cameras. However, the UAV’s payload limitations may prevent it from carrying all sensors simultaneously. \revD{To address this, we are pursuing two complementary approaches.} First, we are designing modular sensor attachments that allow easy reconfiguration of the UAV based on the specific requirements of each inspection mission. Second, Elistair offers a tethered station that includes an optical fiber communication link. In this way, we could remove the computer onboard the UAV, making room for \revD{additional sensors.}

\revD{Another area of interest is the investigation of the applicability of our system in outdoor scenarios, or even in mixed scenarios that encompass both indoor and outdoor areas. These mixed scenarios pose significant challenges for the localization system, as they involve partial GNSS coverage and require ensuring a smooth transition between GNSS and GNSS-denied localization modes. In addition, the environmental factors such as wind can adversely affect the performance of the control system.}

\revD{Finally, we aim to also use this approach in emergency response situations. To this end, we plan to modify the traction type of our UGV platform to enable safe and reliable navigation over uneven terrain.  Additionally, we plan to include a multi-robot SLAM method to perform autonomous exploration of unknown scenarios.}

All the data captured in the experiments is available at \url{https://robotics.upo.es/datasets/marsupial}, accessed on 3 Nov 2025. 

\section*{Acknowledgements}

We would like to thank the staff of the Service Robotics Laboratory at UPO for their unselfish help during the experiments.
We would also like to give special thanks to Sara Victoria Hernández Díaz for suggesting different experimental scenarios for structural inspection and granting us access to them.

This work was partially supported by the grants INSERTION PID2021-127648OB-C31 and RATEC PDC2022-133643-C21, funded by MCIN/AEI/ 10.13039/501100011033 and “European Union NextGenerationEU/PRTR”.


\end{document}